\documentclass[pdflatex,sn-nature]{sn-jnl}%

\usepackage{graphicx}%
\usepackage[percent]{overpic}%
\usepackage{multirow}%
\usepackage{amsmath,amssymb,amsfonts}%
\usepackage{xspace}%
\usepackage{booktabs}%

\begin{document}

\title[EBSD orientation super-resolution]{Symmetry-aware super-resolution of crystal orientation maps via invariant latent-space learning}

\author*[1]{\fnm{Umang} \sur{Garg}}\email{umang@ucsb.edu}
\equalcont{These authors equally led this work}

\author[2]{\fnm{Warren} \sur{Zamudio}}\email{wzamudio@ucsb.edu}
\equalcont{These authors contributed equally to this work.}

\author[3]{\fnm{McLean P.} \sur{Echlin}}\email{mechlin@ucsb.edu}

\author[2]{\fnm{Samantha H.} \sur{Daly}}\email{samdaly@ucsb.edu}

\author[3]{\fnm{Tresa M.} \sur{Pollock}}\email{tresap@ucsb.edu}

\author[1]{\fnm{B.S.} \sur{Manjunath}}\email{manj@ucsb.edu}

\affil*[1]{\orgdiv{Electrical and Computer Engineering}, \orgname{University of California Santa Barbara}, \orgaddress{\city{Santa Barbara}, \postcode{93106}, \state{California}, \country{USA}}}

\affil[2]{\orgdiv{Mechanical Engineering}, \orgname{University of California Santa Barbara}, \orgaddress{\city{Santa Barbara}, \postcode{93106}, \state{California}, \country{USA}}}

\affil[3]{\orgdiv{Materials}, \orgname{University of California Santa Barbara}, \orgaddress{\city{Santa Barbara}, \postcode{93106}, \state{California}, \country{USA}}}

\newcommand{\SO}{\mathrm{SO}(3)}
\newcommand{\RR}{\mathbb{R}}
\newcommand{\FZ}{\mathrm{FZ}}
\newcommand{\Stab}{G}
\newcommand{\TiDataset}{Ti-6Al-4V}
\newcommand{\qrbsaadapted}{Q-RBSA-adapted\xspace}

\abstract{%
Crystal-orientation maps are physical fields defined only up to crystal symmetry; electron backscatter diffraction (EBSD) resolves them experimentally, but acquisition-time constraints limit spatial resolution. Unlike conventional images, EBSD data lie on the quotient space $\SO/\Stab$, where $\Stab$ is the crystal-symmetry group. Standard Euclidean interpolation can therefore mix symmetry-equivalent representations and blur grain boundaries. We introduce the Symmetry-Group-Aware Super-Resolution Attention Network (SG-SRAN), which incorporates crystal symmetry and boundary preservation by design. A frozen, locally isometric encoder maps equivalent orientations to a common latent representation in which Euclidean distance approximates misorientation. Super-resolution is performed in this space, with each high-resolution token restricted to a feature-consistent local support to prevent cross-boundary mixing. A dictionary-based decoder then recovers valid orientations. Across FCC and HCP benchmarks, SG-SRAN matches 15--16-million-parameter backbones using only 27--49k trainable parameters, while achieving the lowest p68 errors, highest inverse-pole-figure fidelity, and zero-shot transfer to unseen alloys.
}

\maketitle

Electron backscatter diffraction (EBSD) is a widely used technique for resolving crystallographic orientation fields in structural materials, with applications ranging from grain-topology characterization to the study of deformation localization, recrystallization and damage initiation~\cite{humphreys2001review}. However, its spatial resolution comes at a substantial acquisition cost. Reducing the pixel pitch increases scan time and limits the area that can be mapped under a fixed measurement budget. A reliable digital super-resolution (SR) method could help relax this trade-off by reconstructing high-resolution microstructural information from more rapidly acquired low-resolution scans (Extended Data Fig.~\ref{fig:ebsd_pipeline}).

\begin{figure}[t]
\centering
\includegraphics[width=\textwidth]{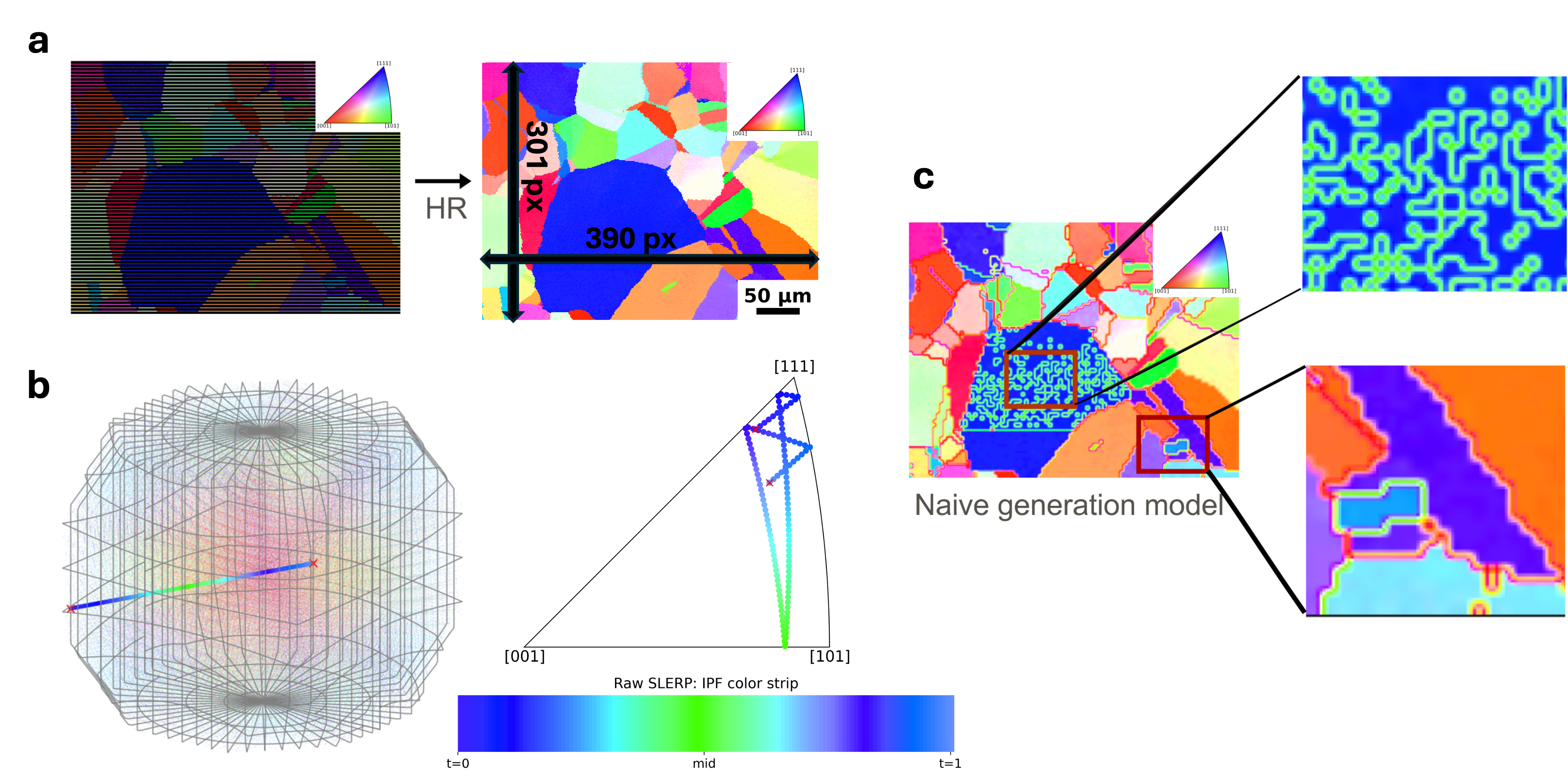}
\caption{\textbf{Geometric and microstructural challenges in orientation-map super-resolution.}
\textbf{a}, A low-resolution (LR) orientation map and its high-resolution (HR) target,
rendered in IPF-$Z$ color. \textbf{b}, Crystal orientations are points on the quotient
manifold $\SO/\Stab$: pairwise spherical-linear interpolation (SLERP) on the quaternion
sphere follows a geodesic but is local and symmetry-blind; the color bar reports the
residual misorientation such an interpolant leaves across an interface.
\textbf{c}, A ``naive'' generative upsampler (SLERP here) that treats orientations as Euclidean image channels; the magnified insets reveal its characteristic, physically invalid failure
modes---speckled intra-grain noise and blurred or bridged grain boundaries---even where the
output looks visually smooth. These artifacts motivate operating in a quotient-aware latent
representation with explicit, boundary-aware routing rather than on rendered RGB
maps. Scale bar, $50\,\mu$m; in the $4\times4$ task, one LR
sample corresponds to a $4\times4$ block of HR pixels.}\label{fig:sr_issue}
\end{figure}

EBSD super-resolution is fundamentally different from conventional image super-resolution because each pixel represents a crystal orientation rather than a Euclidean intensity or color value (Fig.~\ref{fig:sr_issue}a). A crystal orientation is defined only up to the proper rotational symmetry group of the crystal. The physically meaningful orientation space is therefore the quotient manifold $\SO/\Stab$, where $\SO$ denotes the group of three-dimensional rotations and $\Stab$ is the proper rotational crystal point group.

This geometry presents two distinct challenges. First, symmetry-equivalent orientations can have very different numerical representations. Interpolation that neglects crystal symmetry may therefore introduce artificial orientation gradients near fundamental-zone boundaries (Fig.~\ref{fig:sr_issue}b). Second, grain boundaries introduce discontinuities in the physical orientation field. Interpolation or feature aggregation across a grain boundary mixes information from distinct grains, producing nonphysical orientations and blurred, shifted, or smoothed boundaries. A naive upsampler can consequently generate visually smooth reconstructions while introducing speckled intra-grain errors and degraded grain boundaries (Fig.~\ref{fig:sr_issue}c). Common representations, including Euler angles, inverse-pole-figure color maps and raw quaternion coordinates, all retain ambiguities or discontinuities when processed using standard Euclidean operations.

Classical interpolation methods address parts of this problem. For example, Spherical linear interpolation (SLERP) preserves the unit-norm constraint of quaternions ~\cite{shoemake1985slerp}, and symmetry-aware variants can select shorter paths between symmetry-equivalent representations. These approaches nevertheless remain local, pairwise and non-learned. They cannot draw on mesoscale orientation structure or adapt their interpolation support to the surrounding grain topology. Learned EBSD super-resolution has also been explored using IPF-color images and quaternion-valued maps~\cite{seret2019ebsd,jangid2024qrbsa}. Related work has used serial-section learning to generate volumetric EBSD maps~\cite{jangid2024qrbsa}, for which accurate two-dimensional section-level super-resolution is a natural building block. Yet the underlying representation problem remains. When a network operates in a space whose distance metric does not agree with crystallographic misorientation, part of its capacity must be spent compensating for that geometric mismatch before it can learn the reconstruction task.

A learned backbone is also not automatically protected from mixing information across grains. Unless the architecture controls which neighbors contribute to a prediction, feature aggregation can still blur adjacent grains or bridge them with physically incorrect orientation gradients. An effective orientation-field super-resolution method must therefore distinguish between two different situations: orientations that appear different numerically but are equivalent under crystal symmetry, and orientations that are physically distinct because they lie on opposite sides of a grain boundary.

Two developments make this distinction tractable. First, locally isometric embeddings of the quotient space $\SO/\Stab$, constructed from Reynolds-projected Wigner-$D$ features, provide latent coordinates that are invariant to the crystal point-group action while remaining organized into $\SO$ irreducible-representation channels~\cite{hielscher2021locally,arnold2018ambiguous,lenthe2019si}. Symmetry-equivalent orientations therefore map to the same latent representation, while small Euclidean distances approximate small crystallographic misorientations. Second, $\SO$-equivariant operations preserve the transformation laws of these latent features throughout the network rather than flattening them into generic scalar channels~\cite{cohen2016gcnn,weiler2019e2cnn,thomas2018tfn,geiger2022e3nn}.

Here we introduce SG-SRAN, a symmetry-aware and locally routed architecture for orientation-map super-resolution. A frozen encoder maps each input quaternion to a locally isometric latent representation constructed from Reynolds-projected Wigner-$D$ features~\cite{lenthe2019si}. The resulting features are invariant to the proper rotational crystal point group while transforming as structured $\SO$ irreducible-representation channels under global rotations. The trainable SG-SRAN backbone operates entirely on these latent features using $\SO$-equivariant operations implemented in e3nn, preserving their transformation structure throughout the network. Feature-distance-masked latent convolutions restrict local support and reduce information exchange between crystallographically incompatible orientations. During upsampling, neighboring features are grouped into a small set of feature-consistent slots, and each high-resolution (HR) token is routed to its most compatible slot. Finally, a dictionary-based decoder maps the predicted latent features back to unit quaternions using an orientation dictionary generated by cubochoric sampling~\cite{rosca2014cubochoric,rowenhorst2015msmse}.

We evaluate SG-SRAN through three sets of experiments. First, the frozen encoder--dictionary-based decoder reconstructs held-out orientations with milliradian accuracy for both face-centered cubic (FCC) IN718 and hexagonal close-packed (HCP) \TiDataset{} data, showing that the latent space provides a faithful computational interface across distinct crystal symmetries. Second, on the IN718 $4\times4$ benchmark, evaluated over five seeds against four classical interpolants and seven learned baselines, the leading methods are not materially separated by pooled mean misorientation. SG-SRAN nevertheless achieves the lowest p68 and p95 misorientation errors and the highest IPF-space PSNR, while using approximately $320\times$ fewer parameters than the attention-based backbones. This comparison refers to trainable backbone size; end-to-end inference also includes the fixed dictionary and dictionary-based decoder, whose runtime cost is reported in Methods. Third, the same architecture family supports a second crystal symmetry using symmetry-specific invariant encoders and configuration choices. SG-SRAN is the strongest compact model on the \TiDataset{} HCP benchmark under the proper $D_6$ symmetry protocol and supports zero-shot transfer to out-of-distribution FCC CoNi and HCP Ti-Al targets (Table~\ref{tab:zero_shot}). Together, these results show that quotient geometry, equivariant feature processing and local routing provide useful inductive biases for orientation-valued super-resolution, rather than corrections applied after training a conventional Euclidean image backbone.

\section{Results}\label{sec:results}

\subsection{Pipeline overview}

\begin{figure}[h]
\centering
\includegraphics[width=0.97\textwidth]{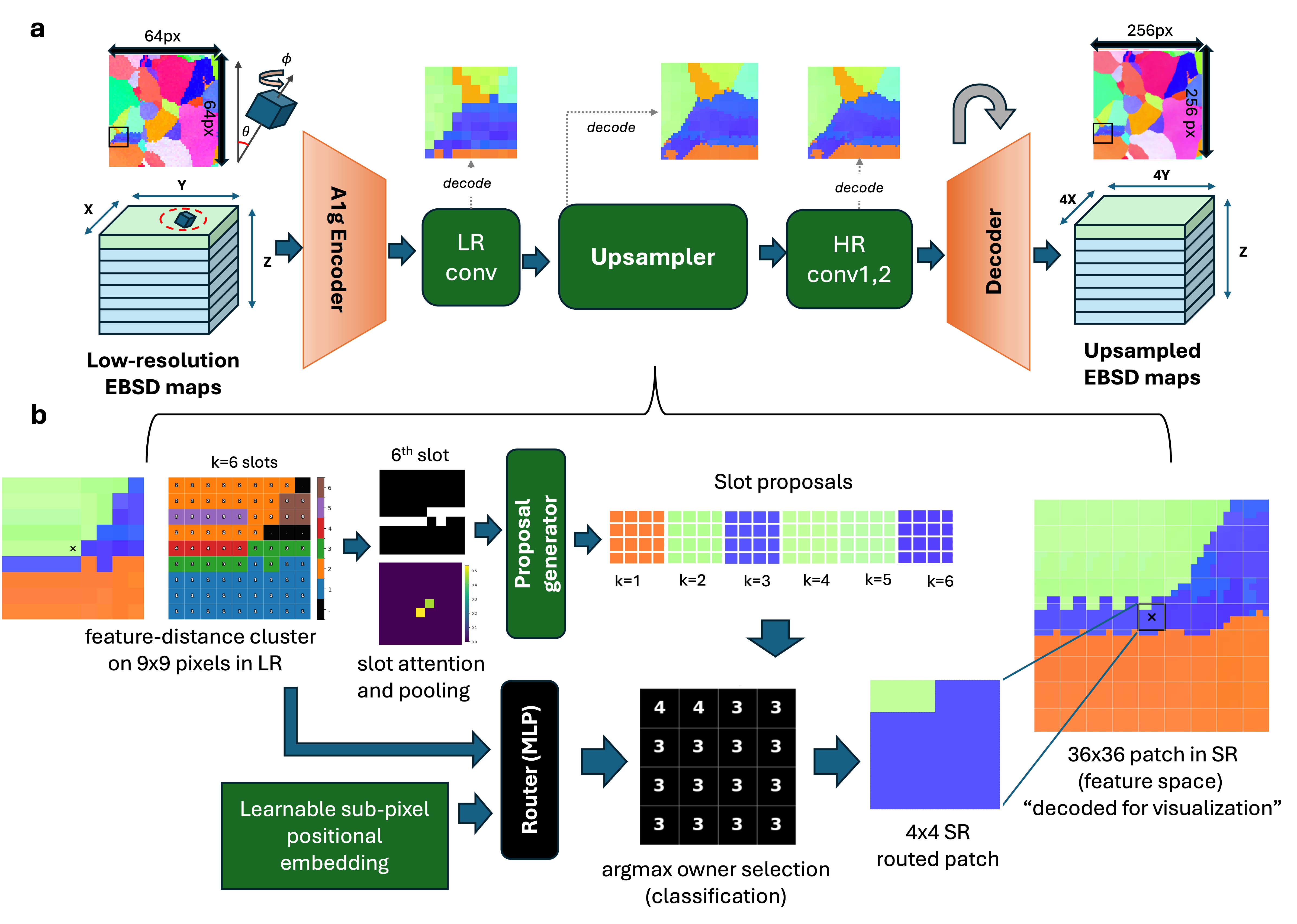}
\caption{%
\textbf{Symmetry-aware latent EBSD super-resolution pipeline and local routed synthesis in SG-SRAN.}
The top row summarizes the end-to-end model. Input orientations are mapped by the frozen $A_{1g}$/Reynolds encoder into a symmetry-invariant latent stack, refined once at LR,
upsampled by SG-SRAN, refined at HR according to the material-specific configuration, and decoded back to crystallographically valid quaternions. The small decoded crops above the latent stacks provide an interpretable view of the orientation field before and after latent upsampling. The lower row expands one local
SG-SRAN decision at a boundary-rich site. The displayed IN718 example uses a $9\times9$ support bank around the marked LR pixel, which is partitioned into candidate orientation slots in the locally isometric latent embedding;
the proposal generator builds token-conditioned candidate features for each slot, while a router uses the complete binary slot-composition masks and learned subpixel position embeddings to assign each HR token to one slot. Each routed HR token is therefore initialized from one observed local orientation branch before HR refinement. Detailed layer definitions and equations
are given in Methods.}%
\label{fig:arch}
\end{figure}

The SG-SRAN pipeline separates the EBSD SR problem into a frozen geometric representation and a
trainable latent-space upsampler (Fig.~\ref{fig:arch}). The encoder collapses
symmetry-equivalent orientations exactly and makes small latent Euclidean
distances correspond to small crystallographic misorientations (see Methods). Therefore, the trainable backbone never compares quaternions directly: local context aggregation, support-bank clustering and
routed patch synthesis all happen in the latent feature space. The routing operation is symmetry-aware: SG-SRAN
clusters a local window-support bank (for instance, the $9\times9$ bank shown in Fig.~\ref{fig:arch}) by Euclidean distance in the locally isometric latent feature
embedding, and the router selects among those symmetry-collapsed feature slots. Thus, upsampling preserves crystal-symmetry
equivalence while discouraging support from distinct grains being merged across a boundary.

Figure~\ref{fig:arch} presents the pipeline at two levels. The top row (a) gives the end-to-end latent SR path: frozen $A_{1g}$ encoding, one LR latent refinement, SG-SRAN upsampling,
configuration-dependent HR latent refinement and decoding. The lower row (b) depicts one typical local SG-SRAN decision: a
boundary-centered local window-support bank is partitioned into latent feature slots,
a proposal generator forms token-conditioned candidate features for each slot, and a MLP router uses the complete binary slot-composition masks together with learned subpixel position to choose the local orientation branch for each HR token in the final $4\times4$ patch. Detailed layer definitions are given in Methods.

\subsection{The frozen encoder--dictionary-based decoder round-trips orientations across FCC and HCP}

\begin{figure*}[t]
\centering
\includegraphics[width=0.96\textwidth]{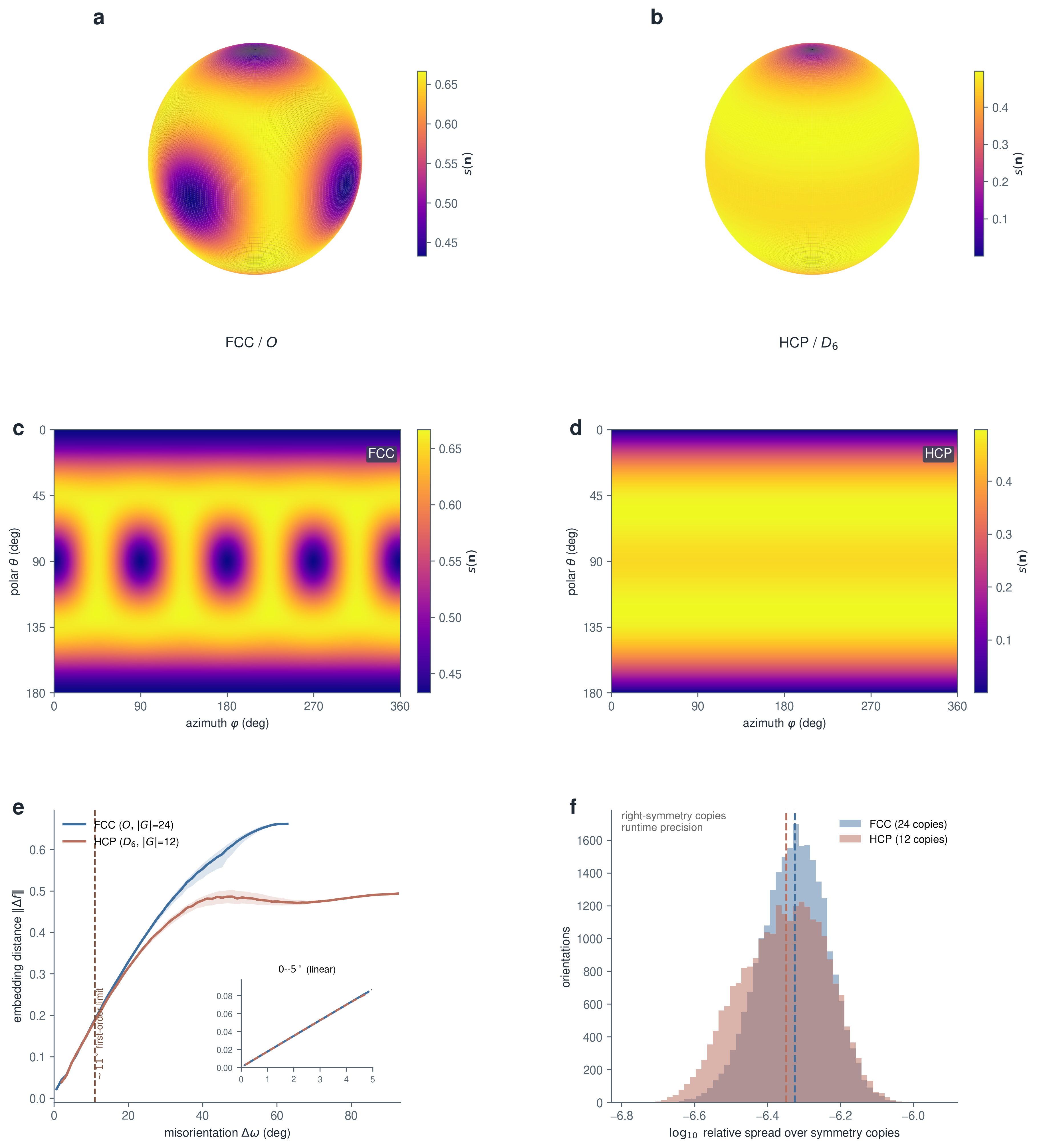}
\caption{%
\textbf{The frozen Reynolds embedding is crystal-symmetry-invariant and locally isometric for both crystal systems.}
\textbf{a},\textbf{b}, Scalar field $s(\mathbf{n})=\lVert f(R(\mathbf{n},60^\circ))-f(I)\rVert$ over
the sphere of rotation axes $\mathbf{n}$, rendered in 3D for the FCC/$O$ (\textbf{a}) and HCP/$D_6$
(\textbf{b}) embeddings; the encoder uses the proper rotational subgroups, while these scalar norm fields also display the corresponding cubic and hexagonal Laue-class visual symmetries ($m\bar{3}m$ and $6/mmm$).
\textbf{c},\textbf{d}, The same fields unrolled on the polar and azimuthal angles $(\theta,\varphi)$,
exposing the symmetry-repeating fundamental tiles (FCC, \textbf{c}; HCP, \textbf{d}).
\textbf{e}, Locally isometric embedding distance $\lVert\Delta f\rVert$ versus symmetry-reduced crystallographic
misorientation $\Delta\omega$ for $2.4\times10^{4}$ low-discrepancy (Fibonacci-lattice sampling) orientation-pair separations per system (line, median;
band, interquartile range). Both systems are linear near the origin (dashed small-angle reference);
the first-order calibration holds to $\sim$0.2~rad ($\sim$$11^\circ$, red dashed) and saturates beyond.
\textbf{f}, Numerical verification of crystal-symmetry invariance. For each low-discrepancy sampled
orientation, we evaluate all $24$ FCC or $12$ HCP symmetry-equivalent copies and report the maximum
deviation from their orbit-mean embedding, normalized by the mean norm. Relative deviations remain
below $1.1\times10^{-6}$, confirming invariance to single-precision accuracy. The retained Reynolds
degrees are $\mathcal{L}=\{4\}$ for FCC and $\mathcal{L}=\{2,4,6\}$ for HCP.%
}\label{fig:encoder}
\end{figure*}

We first isolate the latent interface from the SR task by measuring round-trip
encode--decode fidelity on held-out test orientations: with no spatial upsampling,
how accurately do the frozen encoder and dictionary-based decoder recover the input
orientation? Dense dictionary lookup achieves milliradian-level accuracy on both
crystal systems (Table~\ref{tab:main_results}a): the mean per-sample symmetry-aware
error is $0.0076$~rad on the FCC IN718 test split and $0.0079$~rad on the \TiDataset{} HCP test split. These values measure the end-to-end round-trip fidelity
of the frozen Reynolds encoder and dictionary-based decoder independently of the spatial
super-resolution network. Their close agreement shows that the latent interface attains
comparable round-trip accuracy across the two crystal-symmetry classes. The two geometric
properties that make this latent space a locally faithful interface---exact
crystal-symmetry invariance and local isometry with respect to misorientation---are
verified directly in Fig.~\ref{fig:encoder}; representative round-trip reconstructions
are shown in Extended Data Fig.~\ref{fig:reconstruction}.

\subsection{Benchmark on held-out IN718 and \TiDataset{} at \texorpdfstring{$4\times4$}{4 x 4}}

We evaluate SG-SRAN at scale factor $r=(4,4)$ against four classical interpolants
(nearest-neighbor, bicubic with renormalization, SLERP and symmetry-aware SLERP) and
seven learned baselines trained independently on the matched splits:
\qrbsaadapted~\cite{jangid2024qrbsa}, EDSR~\cite{lim2017edsr}, a quaternion EDSR
(QEDSR)~\cite{lim2017edsr}, an Atindama-style partial-convolution inpainting
network~\cite{atindama2023restoration}, RCAN~\cite{zhang2018rcan},
SAN~\cite{dai2019san} and HAN~\cite{niu2020han}. Every learned
method is trained over five seeds per material and evaluated under a single fixed protocol
(symmetry-aware error $d_\Stab$, a $5^\circ$ boundary mask, the same held-out test splits of
$n=147$ IN718 and $n=72$ \TiDataset{} patches; Methods), so all rows of
Table~\ref{tab:main_results}b are directly comparable. The \TiDataset{} metrics use the
12-element proper $D_6$ rotational subgroup.

On IN718, the attention backbones, the quaternion baselines and SG-SRAN form a closely grouped tier on pooled mean misorientation: RCAN records the lowest mean ($2.52\pm0.04^\circ$), with SG-SRAN ($2.60\pm0.06^\circ$), SAN ($2.62^\circ$) and HAN ($2.63^\circ$) separated by only about a tenth of a degree. The complementary metrics resolve distinct strengths: RCAN leads Boundary F1 ($0.678$) and boundary-band error, QEDSR leads median and interior error, SAN leads boundary-window Composition F1, and SG-SRAN attains the lowest p68 and p95 error percentiles
(p68 $0.81^\circ$ and p95 $1.73^\circ$; the next-best learned p68 is EDSR at
$0.83^\circ$ and the next-best learned p95 is SAN at
$2.70^\circ$, with the remaining learned rows spanning $3.17$--$34.62^\circ$;
Table~\ref{tab:headline}) and the highest IPF-space PSNR and SSIM. SG-SRAN reaches this
with a ${\sim}49$k-parameter trainable backbone---${\sim}320\times$ smaller than RCAN/SAN/HAN,
${\sim}126\times$ smaller than \qrbsaadapted and ${\sim}530\times$ smaller than Atindama
inpainting. The IN718 comparison therefore establishes SG-SRAN's compactness--accuracy advantage:
it leads p68/p95 and parameter efficiency while maintaining competitive accuracy
among other metrics.

On \TiDataset{}, the task is substantially harder because the held-out HCP patches contain a
dense network of fine boundaries. The $27$k-parameter SG-SRAN configuration nevertheless gives the best pooled mean
($9.54\pm0.12^\circ$), median ($0.48^\circ$), interior mean ($0.51^\circ$),
boundary-band mean ($9.63^\circ$), Boundary F1 ($0.639$), PSNR and SSIM. SG-SRAN also leads p68; EDSR leads learned-method p95, and symmetry-aware SLERP leads overall p95 and p99. The Ti-6Al-4V comparison
therefore separates boundary-aware reconstruction from extreme-tail ranking: SG-SRAN gives the
most accurate and most faithful average reconstruction while SAN narrowly gives the best
boundary-window Composition F1 ($0.804$ versus $0.782$ for SG-SRAN).

Qualitative comparisons support the same picture: SG-SRAN and \qrbsaadapted recover coherent
grain interiors while Atindama leaves periodic stripe artifacts from the sparse observation
mask (Extended Data Fig.~\ref{fig:new_learned_baselines}); matched held-out test
examples for both materials are shown in Extended Data Fig.~\ref{fig:visual}. IPF renderings are used only as
qualitative visualizations; quantitative claims rest on $d_\Stab$ and the boundary
diagnostics. Scale dependence across the $2\times2$, $4\times4$ and $8\times8$ tasks on
both materials is summarized in Extended Data Fig.~\ref{fig:scale_dependence}.

\begin{table*}[t]
\caption{\textbf{Model compactness, orientation accuracy and image fidelity (all baselines).} Per-method trainable parameter count (thousands, K), error percentiles (p68/p95/p99) of the per-pixel symmetry-aware misorientation, and IPF-space PSNR/SSIM, on the held-out IN718 ($n=147$) and \TiDataset{} ($n=72$) $4\times4$ test splits. SG-SRAN parameter counts are configuration-specific: 49K for the IN718 support-window setting and 27K for the \TiDataset{} setting. Deterministic interpolants are marked ``--'' and learned rows are five-seed means; $\pm$ reports seed-level standard deviation for learned rows and per-patch test-set standard deviation for deterministic IN718 rows. Pooled mean and median misorientation are in Table~\ref{tab:main_results}\,b. Lower is better for misorientation, higher for PSNR/SSIM; \textbf{bold} marks the best per column. Most IN718 learned methods cluster near $59^\circ$ at p99, whereas the boundary-rich \TiDataset{} benchmark yields larger extreme errors across methods.}\label{tab:headline}
\centering
{\scriptsize
\setlength{\tabcolsep}{4pt}
\resizebox{\textwidth}{!}{%
\begin{tabular}{lcccccc}
\toprule
Method & Params (K) & p68 ($^\circ$) & p95 ($^\circ$) & p99 ($^\circ$) & PSNR (dB) & SSIM \\
\midrule
\multicolumn{7}{l}{\textit{IN718 $4\times4$ (FCC, $n=147$)}} \\
\midrule
Nearest & -- & 1.00 & 46.83 & 59.76 & $17.58 \pm 0.73$ & $0.615 \pm 0.021$ \\
Bicubic & -- & 3.99 & 39.33 & 57.73 & $14.64 \pm 0.90$ & $0.474 \pm 0.044$ \\
SLERP & -- & 1.24 & 38.32 & 56.40 & $14.74 \pm 0.99$ & $0.527 \pm 0.042$ \\
Symm-SLERP & -- & 1.15 & 32.40 & 53.04 & $15.89 \pm 0.65$ & $0.558 \pm 0.031$ \\
\midrule
Atindama inpainting~\cite{atindama2023restoration} & $25{,}784$ & 4.60 & 34.62 & \textbf{50.13} & $13.90 \pm 0.26$ & $0.411 \pm 0.033$ \\
\qrbsaadapted~\cite{jangid2024qrbsa} & $6{,}175$ & 1.03 & 14.64 & 56.23 & $19.01 \pm 0.31$ & $0.652 \pm 0.012$ \\
QEDSR~\cite{lim2017edsr} & $1{,}815$ & 0.86 & 5.23 & 59.21 & $19.91 \pm 0.04$ & $0.706 \pm 0.004$ \\
EDSR~\cite{lim2017edsr} & $7{,}241$ & 0.83 & 3.39 & 59.04 & $20.12 \pm 0.11$ & $0.711 \pm 0.004$ \\
RCAN~\cite{zhang2018rcan} & $15{,}594$ & 0.89 & 3.88 & 58.30 & $20.29 \pm 0.05$ & $0.706 \pm 0.013$ \\
SAN~\cite{dai2019san} & $15{,}862$ & 0.83 & 2.70 & 59.12 & $20.13 \pm 0.06$ & $0.710 \pm 0.004$ \\
HAN~\cite{niu2020han} & $16{,}073$ & 0.85 & 3.17 & 59.11 & $20.10 \pm 0.14$ & $0.709 \pm 0.009$ \\
\textbf{SG-SRAN (ours)} & $\boldsymbol{49}$ & \textbf{0.81} & \textbf{1.73} & 59.55 & $\boldsymbol{20.36 \pm 0.15}$ & $\boldsymbol{0.715 \pm 0.006}$ \\
\midrule
\multicolumn{7}{l}{\textit{\TiDataset{} $4\times4$ (HCP, proper $D_6$, $n=72$)}} \\
\midrule
Nearest & -- & 7.93 & 73.56 & 88.08 & 12.90 & 0.449 \\
Bicubic & -- & 22.45 & 72.26 & 87.23 & 10.59 & 0.262 \\
SLERP & -- & 24.19 & 72.66 & 87.14 & 10.44 & 0.280 \\
Symm-SLERP & -- & 15.58 & \textbf{56.12} & \textbf{79.57} & 12.76 & 0.405 \\
\midrule
Atindama inpainting~\cite{atindama2023restoration} & $25{,}784$ & 22.79 & 69.53 & 86.87 & $10.15 \pm 0.10$ & $0.275 \pm 0.012$ \\
\qrbsaadapted~\cite{jangid2024qrbsa} & $6{,}175$ & 1.93 & 66.00 & 86.61 & $14.29 \pm 0.03$ & $0.557 \pm 0.001$ \\
QEDSR~\cite{lim2017edsr} & $1{,}815$ & 1.99 & 65.58 & 86.49 & $14.36 \pm 0.03$ & $0.562 \pm 0.002$ \\
EDSR~\cite{lim2017edsr} & $7{,}241$ & 6.66 & 61.35 & 84.99 & $13.80 \pm 1.32$ & $0.503 \pm 0.105$ \\
RCAN~\cite{zhang2018rcan} & $15{,}594$ & 3.64 & 63.57 & 85.78 & $14.18 \pm 0.23$ & $0.534 \pm 0.032$ \\
SAN~\cite{dai2019san} & $15{,}862$ & 2.89 & 64.48 & 86.15 & $14.36 \pm 0.05$ & $0.547 \pm 0.009$ \\
HAN~\cite{niu2020han} & $16{,}073$ & 4.00 & 63.34 & 85.79 & $14.22 \pm 0.21$ & $0.529 \pm 0.029$ \\
\textbf{SG-SRAN (ours)} & $\boldsymbol{27}$ & \textbf{1.36} & 65.16 & 86.51 & $\boldsymbol{14.60 \pm 0.06}$ & $\boldsymbol{0.575 \pm 0.005}$ \\
\botrule
\end{tabular}
}
}
\end{table*}

\begin{table*}[t]
\caption{%
\textbf{Encoder fidelity and orientation-accuracy breakdown.}
(\textbf{a}) Frozen Reynolds encoder--dictionary-based decoder round-trip fidelity on held-out test samples, as mean and standard deviation of per-sample mean symmetry-aware error $d_\Stab$ in radians.
(\textbf{b}) IN718 and \TiDataset{} super-resolution accuracy breakdown on the $4\times4$ task: trainable parameter count (K), pooled mean and median per-pixel symmetry-aware misorientation, Boundary F1, boundary-window Composition F1, and the interior / boundary-band split of mean misorientation. SG-SRAN parameter counts are configuration-specific: 49K for IN718 and 27K for \TiDataset{}. Error-tail percentiles and IPF-space PSNR/SSIM are in Table~\ref{tab:headline}. The composition-F1 column is the pooled boundary-window analysis on the selected prediction summaries, with its spurious-rate and recall components reported in Extended Data Table~\ref{tab:boundary_composition_indist}. Lower is better except Boundary F1 and Composition F1; classical interpolants are deterministic and marked ``--'', and the Ti-6Al-4V rows use the 12 proper $D_6$ rotations. Except for the composition-F1 analysis column, learned values are five-seed means and the Mean-column $\pm$ is the seed-level standard deviation. Classical-row $\pm$ values in the Mean column are per-patch test-set standard deviations where available, not seed spreads.
}\label{tab:main_results}
\textbf{a. Encoder--dictionary-based decoder round-trip fidelity}

\vspace{0.3em}
\resizebox{\textwidth}{!}{%
\begin{tabular}{lcc}
\toprule
Variant & IN718 (FCC, $n=147$) & \TiDataset{} (HCP, $n=72$) \\
\midrule
latent features + dictionary lookup & 0.0076 $\pm$ 0.0004 & 0.0079 $\pm$ 0.0001 \\
\botrule
\end{tabular}
}

\vspace{0.8em}
\textbf{b. Orientation-accuracy breakdown on the $4\times4$ task}

\vspace{0.3em}
{\scriptsize
\setlength{\tabcolsep}{3pt}
\resizebox{\textwidth}{!}{%
\begin{tabular}{lcccccc}
\toprule
Method & Params (K) & Mean ($^\circ$) & Median ($^\circ$) & Boundary F1 & Composition F1 & Interior / Boundary mean ($^\circ$) \\
\midrule
\multicolumn{7}{l}{\textit{IN718 $4\times4$ (FCC, $n=147$ held-out test patches; learned methods over five seeds)}} \\
\midrule
Nearest & -- & $4.45 \pm 0.60$ & 1.00 & 0.425 & 0.807 & 0.69 / 9.36 \\
Bicubic & -- & $6.71 \pm 0.99$ & 1.22 & 0.432 & 0.353 & 1.57 / 13.42 \\
SLERP & -- & $6.16 \pm 1.00$ & 0.88 & 0.442 & 0.325 & 1.38 / 12.39 \\
Symm-SLERP & -- & $4.92 \pm 0.65$ & 0.86 & 0.477 & 0.381 & 0.67 / 10.46 \\
\midrule
Atindama inpainting~\cite{atindama2023restoration} & $25{,}784$ & $7.14 \pm 0.42$ & 2.82 & 0.400 & 0.245 & 3.57 / 11.79 \\
\qrbsaadapted~\cite{jangid2024qrbsa} & $6{,}175$ & $3.05 \pm 0.12$ & 0.76 & 0.665 & 0.665 & 0.64 / 6.19 \\
QEDSR~\cite{lim2017edsr} & $1{,}815$ & $2.68 \pm 0.01$ & \textbf{0.60} & 0.626 & 0.853 & \textbf{0.56} / 5.44 \\
EDSR~\cite{lim2017edsr} & $7{,}241$ & $2.59 \pm 0.04$ & 0.61 & 0.647 & 0.849 & 0.58 / 5.22 \\
RCAN~\cite{zhang2018rcan} & $15{,}594$ & $\mathbf{2.52 \pm 0.04}$ & 0.65 & \textbf{0.678} & 0.866 & 0.58 / \textbf{5.05} \\
SAN~\cite{dai2019san} & $15{,}862$ & $2.62 \pm 0.03$ & 0.61 & 0.640 & \textbf{0.899} & 0.58 / 5.27 \\
HAN~\cite{niu2020han} & $16{,}073$ & $2.63 \pm 0.05$ & 0.62 & 0.643 & 0.812 & 0.57 / 5.32 \\
\textbf{SG-SRAN (ours)} & $\boldsymbol{49}$ & $2.60 \pm 0.06$ & 0.61 & 0.658 & 0.891 & 0.61 / 5.20 \\
\midrule
\multicolumn{7}{l}{\textit{\TiDataset{} $4\times4$ (HCP, $n=72$ held-out test patches; learned methods over five seeds)}} \\
\midrule
Nearest & -- & 13.85 & 0.68 & 0.544 & 0.689 & 0.67 / 13.98 \\
Bicubic & -- & 20.96 & 11.68 & 0.611 & 0.293 & 2.71 / 21.14 \\
SLERP & -- & 20.72 & 11.39 & 0.605 & 0.311 & 1.51 / 20.91 \\
Symm-SLERP & -- & 14.94 & 7.22 & 0.561 & 0.406 & 0.73 / 15.08 \\
\midrule
Atindama inpainting~\cite{atindama2023restoration} & $25{,}784$ & $20.76 \pm 0.47$ & 13.03 & 0.618 & 0.263 & 8.50 / 20.88 \\
\qrbsaadapted~\cite{jangid2024qrbsa} & $6{,}175$ & $10.36 \pm 0.09$ & 1.00 & 0.620 & 0.773 & 0.92 / 10.46 \\
QEDSR~\cite{lim2017edsr} & $1{,}815$ & $10.09 \pm 0.03$ & 0.66 & 0.619 & 0.766 & 0.59 / 10.18 \\
EDSR~\cite{lim2017edsr} & $7{,}241$ & $11.72 \pm 3.55$ & 3.34 & 0.609 & 0.739 & 2.72 / 11.81 \\
RCAN~\cite{zhang2018rcan} & $15{,}594$ & $10.74 \pm 0.69$ & 1.88 & 0.613 & 0.511 & 1.27 / 10.83 \\
SAN~\cite{dai2019san} & $15{,}862$ & $10.27 \pm 0.15$ & 1.18 & 0.621 & \textbf{0.804} & 0.95 / 10.36 \\
HAN~\cite{niu2020han} & $16{,}073$ & $10.64 \pm 0.64$ & 1.79 & 0.624 & 0.671 & 1.26 / 10.73 \\
\textbf{SG-SRAN (ours)} & $\boldsymbol{27}$ & $\boldsymbol{9.54 \pm 0.12}$ & \textbf{0.48} & \textbf{0.639} & 0.782 & \textbf{0.51} / \textbf{9.63} \\
\botrule
\end{tabular}
}
}

\end{table*}

\subsection{Grain Boundary analysis}

The boundary diagnostics measure complementary reconstruction properties: strict
Boundary F1 scores edge localization, boundary-band misorientation scores angular
accuracy, and boundary-window Composition F1 scores recovery of the distinct local
orientation branches present within an interface window. RCAN leads IN718 Boundary F1,
while SG-SRAN leads the IN718 p68/p95 errors and the \TiDataset{} mean, median, and
boundary-band metrics. Table~\ref{tab:main_results}b reports Composition F1, and
Extended Data Table~\ref{tab:boundary_composition_indist} separates its spurious-rate
and recall components. Spurious rate is the fraction of SR observations in boundary windows that do not match any HR orientation group within the prescribed misorientation threshold. Nearest-neighbor copying produces few spurious observations
but misses minority local orientation branches; SG-SRAN reaches $0.891$ on IN718 and
$0.782$ on \TiDataset{}, close to SAN's respective leading values of $0.899$ and
$0.804$.

The evidence for the mechanism is therefore local, matching the scale at which
SG-SRAN forms routed patch assignments. The walkthrough in Extended Data
Fig.~\ref{fig:probe_routing} shows that crop-level owner changes and reduced router
confidence follow grain-boundary neighborhoods, while support-bank clustering and
decoded candidate features remain consistent with the local orientation branches in the same
locally isometric latent feature space. This supports the interpretation that SG-SRAN
selects among local orientation branches rather than smoothing across the interface.
These owner, confidence, and slot-usage maps also make the learned upsampler
inspectable: failure modes can be examined in the same latent feature space used for
routing, rather than only through the final decoded IPF maps.

The stagewise decoded-error analysis provides a complementary aggregate check on both
crystal systems. The routed upsampler accounts for nearly all of the reduction in
decoded error from the encoded input to the final output: $98.0\%$ on IN718
(147 held-out test samples) and $99.7\%$ on \TiDataset{}
(72 held-out test samples), with subsequent HR refinement changing the mean error by
at most $0.02^\circ$ in either material
(Extended Data Fig.~\ref{fig:stagewise_progression}). The no-routing control isolates
the contribution of the routed-patch upsampler: replacing only this component with
latent bicubic or nearest-neighbor interpolation, while holding every other component
fixed, reduces IN718 Boundary F1 to $0.445$ and $0.425$, respectively, compared with
$0.656$ for the full pipeline (Methods).

\subsection{Zero-shot transfer to out-of-distribution FCC and HCP datasets}
Because the pipeline is built around a frozen symmetry-aware representation, trained
checkpoints can be applied unchanged to new materials within the same symmetry class.
Table~\ref{tab:zero_shot} reports zero-shot transfer to two out-of-distribution targets: a CoNi alloy (FCC) and a Ti-Al alloy (HCP).
SG-SRAN remains strongest on CoNi mean, median, p68, p95 and IPF fidelity, while HAN
has the highest CoNi Boundary F1 and SAN has the highest CoNi Composition F1. On Ti-Al,
the Ti-6Al-4V-trained SG-SRAN gives the best mean, boundary-band mean, Boundary F1, composition
F1, p68, p95, p99, PSNR and SSIM among the reported methods.
The zero-shot boundary-Composition F1 values are integrated into Table~\ref{tab:zero_shot}a, with the full spurious-rate and recall decomposition in Extended Data Table~\ref{tab:zero_shot_boundary_composition}. This analysis separates the two transfer regimes: SAN narrowly gives the highest CoNi Composition F1 ($0.844$ versus $0.834$ for SG-SRAN), whereas SG-SRAN gives the highest Ti-Al Composition F1 ($0.915$).
Representative IPF-$Z$ maps for two CoNi and two Ti-Al zero-shot samples, including all deterministic and learned baselines, are shown in Extended Data Fig.~\ref{fig:zero_shot_ipfz_all_methods}.

\begin{table*}[p]
\caption{%
\textbf{Zero-shot transfer to out-of-distribution FCC and HCP targets.}
All entries use source-domain checkpoints without target retraining. Composition F1 is the
boundary-window composition score; its spurious-rate and recall components are reported in
Extended Data Table~\ref{tab:zero_shot_boundary_composition}. Lower is better for angular
metrics; higher is better for Boundary F1, Composition F1, PSNR and SSIM. \textbf{Bold} marks
the best available method in each column. SG-SRAN uses 49K trainable parameters for the
IN718-source FCC transfer and 27K for the \TiDataset{}-source HCP transfer.}
\label{tab:zero_shot}
\centering
\textbf{a. Orientation-accuracy and boundary metrics.}
\vspace{0.3em}
{\scriptsize
\setlength{\tabcolsep}{3.0pt}
\resizebox{0.99\textwidth}{!}{%
\begin{tabular}{llcccccc}
\toprule
Method & Params (K) & Mean ($^\circ$) & Median ($^\circ$) & Boundary F1 & Composition F1 & Interior ($^\circ$) & Boundary band ($^\circ$) \\
\midrule
\multicolumn{8}{l}{\textit{IN718 $\to$ CoNi (FCC): zero-shot test split $n=20$}} \\
\midrule
Nearest & -- & 2.85 & 0.27 & 0.405 & 0.741 & 0.35 & 8.30 \\
Bicubic & -- & 3.94 & 0.40 & 0.446 & 0.387 & 0.47 & 11.53 \\
SLERP & -- & 3.49 & 0.27 & 0.452 & 0.364 & 0.33 & 10.40 \\
Symm-SLERP & -- & 3.12 & 0.27 & 0.455 & 0.401 & 0.32 & 9.22 \\
\midrule
Atindama inpainting~\cite{atindama2023restoration} & $25{,}784$ & 7.87 & 2.49 & 0.260 & 0.240 & 5.80 & 12.40 \\
\qrbsaadapted~\cite{jangid2024qrbsa} & $6{,}175$ & 2.35 & 0.88 & 0.666 & 0.636 & 0.91 & 5.49 \\
QEDSR~\cite{lim2017edsr} & $1{,}815$ & 1.66 & 0.23 & 0.628 & 0.801 & 0.29 & 4.65 \\
EDSR~\cite{lim2017edsr} & $7{,}241$ & 1.65 & 0.25 & 0.644 & 0.798 & 0.30 & 4.59 \\
RCAN~\cite{zhang2018rcan} & $15{,}594$ & 1.66 & 0.24 & 0.640 & 0.804 & 0.29 & 4.65 \\
SAN~\cite{dai2019san} & $15{,}862$ & 1.62 & 0.26 & 0.631 & \textbf{0.844} & 0.31 & 4.48 \\
HAN~\cite{niu2020han} & $16{,}073$ & 1.68 & 0.32 & \textbf{0.675} & 0.760 & 0.34 & 4.61 \\
\textbf{SG-SRAN (ours)} & $\boldsymbol{49}$ & \textbf{1.59} & \textbf{0.23} & 0.670 & 0.834 & \textbf{0.29} & \textbf{4.42} \\
\midrule
\multicolumn{8}{l}{\textit{\TiDataset{} $\to$ Ti-Al (HCP): zero-shot test split $n=133$}} \\
\midrule
Nearest & -- & 2.21 & 0.36 & 0.393 & 0.818 & 0.38 & 7.55 \\
Bicubic & -- & 4.08 & 0.43 & 0.364 & 0.395 & 0.66 & 14.04 \\
SLERP & -- & 3.78 & 0.33 & 0.404 & 0.411 & 0.53 & 13.26 \\
Symm-SLERP & -- & 2.48 & 0.33 & 0.415 & 0.462 & 0.35 & 8.68 \\
\midrule
Atindama inpainting~\cite{atindama2023restoration} & $25{,}784$ & 12.01 & 6.47 & 0.115 & 0.243 & 9.79 & 18.51 \\
\qrbsaadapted~\cite{jangid2024qrbsa} & $6{,}175$ & 2.05 & 0.79 & 0.533 & 0.867 & 0.80 & 5.69 \\
QEDSR~\cite{lim2017edsr} & $1{,}815$ & 1.68 & 0.38 & 0.536 & 0.850 & 0.42 & 5.37 \\
EDSR~\cite{lim2017edsr} & $7{,}241$ & 1.72 & 0.43 & 0.554 & 0.823 & 0.48 & 5.36 \\
RCAN~\cite{zhang2018rcan} & $15{,}594$ & 5.15 & 2.72 & 0.456 & 0.456 & 3.76 & 9.20 \\
SAN~\cite{dai2019san} & $15{,}862$ & 1.99 & 0.67 & 0.514 & 0.900 & 0.76 & 5.60 \\
HAN~\cite{niu2020han} & $16{,}073$ & 3.07 & 1.45 & 0.470 & 0.711 & 1.85 & 6.62 \\
\textbf{SG-SRAN (ours)} & $\boldsymbol{27}$ & \textbf{1.21} & \textbf{0.30} & \textbf{0.630} & \textbf{0.915} & \textbf{0.33} & \textbf{3.78} \\
\botrule
\end{tabular}
}
}
\end{table*}

\begin{table*}[p]
\centering
\textbf{Table~\ref{tab:zero_shot}b. Error-tail percentiles and IPF image fidelity.}
\vspace{0.3em}
{\scriptsize
\setlength{\tabcolsep}{3.0pt}
\resizebox{0.99\textwidth}{!}{%
\begin{tabular}{llccccc}
\toprule
Method & Params (K) & p68 ($^\circ$) & p95 ($^\circ$) & p99 ($^\circ$) & PSNR (dB) & SSIM \\
\midrule
\multicolumn{7}{l}{\textit{IN718 $\to$ CoNi (FCC): zero-shot test split $n=20$}} \\
\midrule
Nearest & -- & 0.42 & 24.04 & 59.76 & 19.69 & 0.770 \\
Bicubic & -- & 0.95 & 26.12 & 52.14 & 17.61 & 0.680 \\
SLERP & -- & 0.46 & 26.42 & 50.57 & 17.89 & 0.728 \\
Symm-SLERP & -- & 0.46 & 22.37 & 47.61 & 18.36 & 0.733 \\
\midrule
Atindama inpainting~\cite{atindama2023restoration} & $25{,}784$ & 5.47 & 40.48 & 48.59 & 14.03 & 0.402 \\
\qrbsaadapted~\cite{jangid2024qrbsa} & $6{,}175$ & 1.16 & 5.62 & \textbf{46.23} & 21.04 & 0.779 \\
QEDSR~\cite{lim2017edsr} & $1{,}815$ & 0.34 & 1.62 & 53.84 & 22.28 & 0.831 \\
EDSR~\cite{lim2017edsr} & $7{,}241$ & 0.37 & 1.67 & 53.05 & 22.30 & 0.831 \\
RCAN~\cite{zhang2018rcan} & $15{,}594$ & 0.36 & 1.62 & 53.69 & 22.25 & 0.828 \\
SAN~\cite{dai2019san} & $15{,}862$ & 0.37 & 1.43 & 54.71 & 22.51 & 0.837 \\
HAN~\cite{niu2020han} & $16{,}073$ & 0.49 & 2.64 & 50.14 & 22.44 & 0.813 \\
\textbf{SG-SRAN (ours)} & $\boldsymbol{49}$ & \textbf{0.32} & \textbf{1.20} & 54.97 & \textbf{22.66} & \textbf{0.842} \\
\midrule
\multicolumn{7}{l}{\textit{\TiDataset{} $\to$ Ti-Al (HCP): zero-shot test split $n=133$}} \\
\midrule
Nearest & -- & 0.49 & 1.31 & 64.92 & 22.15 & 0.828 \\
Bicubic & -- & 0.72 & 22.50 & 71.33 & 18.75 & 0.732 \\
SLERP & -- & 0.48 & 25.55 & 73.02 & 18.81 & 0.771 \\
Symm-SLERP & -- & 0.47 & 16.13 & 46.46 & 21.50 & 0.814 \\
\midrule
Atindama inpainting~\cite{atindama2023restoration} & $25{,}784$ & 11.10 & 51.19 & 80.47 & 13.49 & 0.334 \\
\qrbsaadapted~\cite{jangid2024qrbsa} & $6{,}175$ & 0.96 & 1.60 & 56.70 & 23.48 & 0.828 \\
QEDSR~\cite{lim2017edsr} & $1{,}815$ & 0.53 & 1.44 & 54.82 & 23.69 & 0.833 \\
EDSR~\cite{lim2017edsr} & $7{,}241$ & 0.59 & 1.67 & 51.04 & 23.49 & 0.824 \\
RCAN~\cite{zhang2018rcan} & $15{,}594$ & 3.79 & 17.65 & 50.42 & 18.25 & 0.599 \\
SAN~\cite{dai2019san} & $15{,}862$ & 0.89 & 2.64 & 54.47 & 23.31 & 0.763 \\
HAN~\cite{niu2020han} & $16{,}073$ & 2.16 & 6.71 & 47.56 & 21.05 & 0.687 \\
\textbf{SG-SRAN (ours)} & $\boldsymbol{27}$ & \textbf{0.40} & \textbf{0.88} & \textbf{42.61} & \textbf{25.31} & \textbf{0.886} \\
\botrule
\end{tabular}
}
}
\end{table*}

\clearpage

\section{Discussion}\label{sec:discussion}

Orientation super-resolution should be treated as reconstruction on the symmetry quotient
$\SO/G$, rather than as interpolation of generic image channels. By combining this geometry
with local routing at grain boundaries, SG-SRAN produces accurate, physically valid outputs
through an inspectable routed upsampling process. Across five seeds, the compact
27--49k-parameter model performs strongly on tail-error and boundary-sensitive metrics across
both crystal systems, although competing methods lead individual measures. Orientation-field
super-resolution should therefore be assessed jointly through angular accuracy, boundary
localization, local orientation-branch recovery, and compactness.

The boundary performance arises from the geometry-aware representation and routed upsampler
acting together. The right-$G$-invariant, locally isometric encoder gives nearby Euclidean
latent distances a first-order interpretation as crystallographic misorientation. Routing
changes and reduced confidence concentrate near grain boundaries, where the upsampler forms
distinct candidate features and assigns them to the high-resolution output grid. The slot assignments are made during upsampling; subsequent convolution layers refine the latent features without re-clustering the local support, while boundary-window Composition F1 measures whether the
corresponding local branches survive in the final output. SG-SRAN therefore preserves interfaces
by assigning among local orientation branches rather than averaging across them.

The principal limitations are difficult orientation selection, decoding cost, and validation
breadth. On the boundary-rich \TiDataset{} benchmark, the most extreme p95--p99 errors remain
comparable to those of other learned methods; because dictionary-based decoding always returns a
canonical unit quaternion, these errors are valid but incorrect orientation choices rather
than invalid states. Dictionary search and refinement dominate inference, requiring
approximately $4.63$~s per patch compared with $28.5$~ms for encoding and latent
super-resolution. The CoNi and Ti--Al transfers provide initial evidence within FCC and HCP
symmetry classes, but broader cross-specimen validation remains necessary.

These findings point directly to router calibration, learnable decoding, and
broader cross-specimen validation. More broadly, they support a representation-first strategy:
quotient geometry defines physical equivalence, local-isometry normalization gives nearby
latent distances a crystallographic meaning, and routed upsampling preserves distinct local
branches at interfaces. Building these properties into both the representation and the
upsampling process, rather than correcting a Euclidean image model after the fact, may provide
a useful strategy for reconstructing fields defined only up to a physical symmetry.

\section{Methods}\label{sec:methods}

\paragraph{Global notation.}
We write $Q_\mathrm{LR}$ and $Q_\mathrm{HR}$ for the low- and high-resolution quaternion
fields, $E$ and $D$ for the frozen encoder and dictionary-based decoder, and $d_\Stab$ for the symmetry-aware misorientation metric of equation~\ref{eq:sym_err}. The spatial upsampling factor is
$r=(r_y,r_x)$ and the task studied here is $r=(4,4)$. Latent feature fields after successive backbone stages are denoted $Z_0,\dots,Z_4$ as in equation~\ref{eq:ocrp_pipeline}.

\subsection{Crystal orientations and quotient geometry}
In this work, crystal orientation is represented as an active crystal-to-specimen rotation $R\in\SO$, defined by $v_s=Rv_c$, where $v_c$ and $v_s$ are the coordinates of the same physical direction in right-handed crystal and specimen frames, respectively. We write $R=R(q)$ and represent it by a scalar-first Hamilton unit quaternion $q=(q_0,q_1,q_2,q_3)\in S^3$, acting as $R(q)v=\operatorname{vec}\left(q\otimes(0,v)\otimes q^{-1}\right)$, where $(0,v)$ denotes the pure quaternion with zero scalar part and vector part $v$, and $\operatorname{vec}(\cdot)$ extracts the vector part of a quaternion. Under this convention, composition satisfies $R(q_1\otimes q_2)=R(q_1)R(q_2)$. The antipodal quaternions $q$ and $-q$ represent the same rotation, so $\SO\simeq S^3/{\pm1}$. When an EBSD tool supplies a passive specimen-to-crystal orientation, we denote it by $R_{cs}=R(q_{cs})$; after expressing it in our crystal and specimen frames, we convert it once at import as $R=R_{cs}^{-1}=R_{cs}^{\mathsf T}$ and $q=q_{cs}^{-1}$. Thereafter, $R$ and $q$ always denote active crystal-to-specimen orientations, with conversion back to the source convention performed only for visualization. Our frame, handedness, and active/passive conventions follow Britton et al.~\cite{britton2016whichway}.

The crystal lattice has a finite proper rotational crystal point group $G\subset\SO$, whose elements act actively in the crystal frame. For each $g\in G$, let $q_g\in S^3$ be a unit-quaternion lift satisfying $R(q_g)=g$. Because $R$ maps crystal coordinates to specimen coordinates, crystal symmetry acts from the right: $R\sim Rg$, equivalently $q\sim q\otimes q_g$. Thus, the physical orientation space is the quotient manifold $\SO/G$. For face-centered cubic (FCC) metals, $G$ is the cubic rotation group $O$ with $|O|=24$; for hexagonal close-packed (HCP) metals, it is the dihedral rotation group $D_6$ with $|D_6|=12$.

The fundamental zone (FZ) selects one representative from each symmetry orbit, up to boundary ties. Enumerating
$G=\{g_i\}_{i=1}^{|G|}$, we form
\begin{equation}
\begin{aligned}
q^{(i)} &= q\otimes q_{g_i}, \qquad
i^\star = \arg\max_i \left|w^{(i)}\right|,
\end{aligned}
  \label{eq:fz_reduce}
\end{equation}
where $w^{(i)}$ is the scalar part of $q^{(i)}$. We define
$q_{\mathrm{FZ}}=q^{(i^\star)}$ and choose its antipodal sign such that
$w_{\mathrm{FZ}}\geq 0$. This selects the symmetry-equivalent rotation closest to the identity; remaining boundary ties are resolved deterministically. The symmetry-aware angular error is
\begin{equation}
d_G(q_1,q_2)
=
\min_{g\in G}
2\arccos\left(
\left|
\left\langle
q_1,\,
q_2\otimes q_g
\right\rangle
\right|
\right),
  \label{eq:sym_err}
\end{equation}
which is zero for symmetry-equivalent orientations and reduces to
$
d_{\mathrm{geo}}(q_1,q_2)
=
2\arccos\left|
\left\langle q_1,q_2\right\rangle
\right|
$
when $|G|=1$.

\subsection{Locally isometric orientation encoder}

The encoder maps each pixel quaternion to a finite-dimensional Euclidean vector that is
(i)~exactly invariant under $G$ and (ii)~locally isometric such that small latent distances
track small geodesic distances on $\SO/G$ near the identity equivalence class.

\paragraph{Symmetry-invariant feature blocks.}
For each angular degree $l$, the real (tesseral) Wigner matrix
$D^l : \SO \to \RR^{(2l+1)\times(2l+1)}$ is the degree-$l$ irreducible representation of $\SO$,
obtained from the complex Wigner-$D$ matrices by the unitary change of basis to real tesseral
harmonics; it preserves the homomorphism $D^l(R_1 R_2) = D^l(R_1)D^l(R_2)$ and keeps all
features and weights real. The Reynolds projector
\begin{equation}
  P_l^{(G)} := \frac{1}{|G|}\sum_{g \in G} D^l(g)
  \label{eq:reynolds}
\end{equation}
projects onto the $G$-fixed subspace
$I_l(G) = \{u : D^l(g)u = u \ \forall g \in G\}$.
With $r_l = \dim I_l(G)$ and
$U_l^{(G)} \in \RR^{(2l+1)\times r_l}$ an orthonormal basis of $I_l(G)$,
the crystal-invariant feature block at degree $l$ is
\begin{equation}
  F_l(q) := D^l(R(q))\,U_l^{(G)} \in \RR^{(2l+1) \times r_l},
  \label{eq:block}
\end{equation}
where $R(q)$ is the active rotation of $q$. Because
$D^l(g)\,U_l^{(G)} = U_l^{(G)}$ for all $g \in G$,
invariance under $R(q)\sim R(q)g$ follows from the homomorphism property and
$R(q \otimes q_g) = R(q)g$:
\begin{equation}
  F_l(q \otimes q_g)
  = D^l\!\bigl(R(q)g\bigr)\,U_l^{(G)}
  = D^l(R(q))\,\underbrace{D^l(g)\,U_l^{(G)}}_{=\,U_l^{(G)}}
  = F_l(q).
  \label{eq:invariance}
\end{equation}
Two group actions must be distinguished: the encoder is \emph{invariant} under the right
crystal-symmetry action $q\mapsto q\otimes q_g$
(equation~\ref{eq:invariance}), which collapses symmetry-equivalent orientations, and
\emph{equivariant} under the left global-rotation action
$q\mapsto q'\otimes q$, with $R'=R(q')$ (i.e.\ $R\mapsto R'R$), for which
$F_l(q)\mapsto D^l(R')\,F_l(q)$. The backbone operates on these right-$G$-invariant feature blocks as
left-$\SO$-equivariant e3nn representations---processing orientation as
$\SO$ tensors ---while the routing and
feature-matching loss introduced in Section $3.3$ are constructed from invariant contractions of
these equivariant features. The raw descriptor is the concatenation
$\Phi_\mathrm{raw}(q)
= \bigoplus_{l \in \mathcal{L}} \mathrm{vec}(F_l(q))$
over the selected invariant degrees $\mathcal{L}$.

\paragraph{Constructing the invariant bases.}
The invariant bases $U_l^{(\Stab)}$ are built once, offline, and stored as non-trainable
buffers. For each degree $l$ in the truncation set, we form the Reynolds projector of
equation~\ref{eq:reynolds} in the real tesseral Wigner basis, symmetrize it numerically, and take the eigenvectors with eigenvalue one as an orthonormal basis
$U_l^{(\Stab)}$ for the fixed subspace. The scalar block $l=0$ is mathematically invariant but is
not emitted by the orientation encoder because it carries no orientation information. The emitted
descriptor therefore contains only non-scalar invariant blocks.

For FCC ($O$), the descriptor is truncated at $l\le4$. Since $r_2=0$ and $r_4=1$, the first
non-trivial cubic invariant is the $9$-dimensional $1\!\times\!4e$ block. For HCP ($D_6$), the
descriptor is truncated at $l\le6$ and keeps every non-scalar projector component below that
cutoff: $r_2=1$, $r_4=1$, and $r_6=2$. The HCP encoder output is therefore
$1\!\times\!2e + 1\!\times\!4e + 2\!\times\!6e$, i.e.\ $5+9+2\times13=40$ dimensions, and serves
as both the encoder output and the backbone working space.

\paragraph{Local-isometry normalization.}
The Reynolds projector fixes the invariant subspace $U_l^{(G)}$, but it does not fix
the Euclidean scale assigned to each emitted feature block. We therefore calibrate the scale of
the actual encoder feature map at the identity. For each retained block
$F_l(q)=D^l(R(q))U_l^{(G)}$, let
$J_l\in\RR^{d_l\times3}$, with $d_l=(2l+1)r_l$, denote the Jacobian of
$\mathrm{vec}(F_l(q))$ with respect to infinitesimal axis-angle perturbations at
$R(q)=\mathrm{Id}$. The normalized encoder is
\[
  E(q)=\bigoplus_{l\in\mathcal{L}}
  \beta_l\,\mathrm{vec}(F_l(q)),
\]
where the scalar block weights $\beta_l$ are chosen so that
\begin{equation}
  \sum_{l\in\mathcal{L}}
  \beta_l^2 J_l^\top J_l
  \approx I_3.
  \label{eq:block_iso}
\end{equation}
These scalar weights change only the metric scale; they do not change the invariant subspaces or
the right-$G$ invariance established in equation~\ref{eq:invariance}.

For FCC ($G=O$), only the $1\times4e$ block is retained through $l\leq4$. Its unnormalized
local metric is $J_4^\top J_4=(20/3)I_3$ to numerical precision, so we use
\[
  \beta_4=\sqrt{\frac{3}{20}}=0.3872983346.
\]
For HCP ($G=D_6$), the retained blocks through $l\leq6$ are
$1\times2e+1\times4e+2\times6e$. Their unnormalized local metric contributions are $
  J_2^\top J_2\approx\operatorname{diag}(3,3,0),$
  $J_4^\top J_4\approx\operatorname{diag}(10,10,0)$,
  $J_6^\top J_6\approx\operatorname{diag}(24,24,36)$.
With equal weights, the combined local metric would be
$\operatorname{diag}(37,37,36)$, corresponding to the small anisotropy
$\lambda_{\max}/\lambda_{\min}=37/36\approx1.03$. The implemented normalization instead uses
\[
  \beta_2=0.1633674001,\qquad
  \beta_4=0.1591435236,\qquad
  \beta_6=0.1666666667,
\]
which makes the combined local metric equal to $I_3$ up to numerical precision.

Thus, for a sufficiently small axis-angle vector $\xi\in\RR^3$, let
$\widehat{\xi}\in\mathfrak{so}(3)$ denote its skew-symmetric matrix, and let
$q_\xi\in S^3$ be a unit-quaternion lift satisfying
$R(q_\xi)=\exp(\widehat{\xi})$. Let
$q_{\mathrm{Id}}=(1,0,0,0)$, so that $R(q_{\mathrm{Id}})=I_3$. Then
\begin{equation}
  \left\|E(q_\xi)-E(q_{\mathrm{Id}})\right\|
  =
  \|\xi\|+O(\|\xi\|^2).
  \label{eq:local_iso}
\end{equation}
Consequently, mean-squared error in the normalized latent space tracks squared geodesic
orientation error to leading order. This local-isometry construction follows
\cite{hielscher2021locally}. In our numerical analysis, the approximation remains tight for
misorientations up to approximately $0.2$~rad, or $11^\circ$, beyond which higher-order
Taylor terms become non-negligible.

\paragraph{Completeness and truncation.}
In the untruncated harmonic construction, all nonconstant degrees with
$r_l=\dim I_l(G)>0$ together separate quotient orientations in $\SO/G$. The implemented
encoders retain finite degree sets, $\mathcal{L}=\{4\}$ for FCC and
$\mathcal{L}=\{2,4,6\}$ for HCP, yielding finite-dimensional representations. The selected
degrees determine the harmonic content, latent dimension and computational cost of each
encoder. The local-isometry analysis characterizes the infinitesimal geometry of these
representations, while the encoder--dictionary round-trip error in
Table~\ref{tab:main_results} measures orientation recovery by the implemented
encoder--dictionary pair on the orientations used in the experiments.

\subsection{Orientation-cluster routed-patch super-resolution backbone}

The SG-SRAN super-resolution stage uses an orientation-cluster routed-patch upsampler that acts
directly on the locally isometric latent field, predicting one high-resolution (HR) latent patch
per low-resolution (LR) pixel from a local, orientation-conditioned support bank.
The emitted patch is $4\times4$ for the $r=(4,4)$ SR task. Every stage operates on the crystal-invariant
latent of the encoder (the $9$-dimensional $l=4$ block for FCC, the $40$-dimensional
$1\!\times\!2e+1\!\times\!4e+2\!\times\!6e$ descriptor for HCP), so crystal invariance of the predicted orientation is
preserved end to end.

Let $Z_0 = E(Q_\mathrm{LR}) \in \RR^{B \times HW \times C}$ be the encoded LR feature map, with
batch size $B$, flattened LR grid $H \times W$, and latent dimension $C$. The pipeline is
\begin{equation}
  Z_1 = \mathcal{C}^{\mathrm{LR}}_{w_\mathrm{LR}}(Z_0), \quad
  Z_2 = \mathcal{O}(Z_1;\,Z_0), \quad
  Z_3 = \mathcal{C}^{\mathrm{HR}}_{w_\mathrm{HR}}(Z_2), \quad
  Z_4 = \mathcal{R}^{\mathrm{HR}}(Z_3),
  \label{eq:ocrp_pipeline}
\end{equation}
where $\mathcal{C}^{\mathrm{LR}}_{w_\mathrm{LR}}$ and $\mathcal{C}^{\mathrm{HR}}_{w_\mathrm{HR}}$
are feature-similarity-masked equivariant convolutions, $\mathcal{O}$ is the SG-SRAN upsampler
with $Z_0$ used for clustering and $Z_1$ used for token synthesis,
and $\mathcal{R}^{\mathrm{HR}}$ denotes the optional second HR refinement pass. SG-SRAN uses a
configured odd local support window (e.g. $9\times9$ for IN718 and $5\times5$ for \TiDataset{}),
up to $K=6$ candidate slots per LR pixel, and an 8-neighbor feature-space
clustering graph. The router is a small MLP ($\mathtt{hidden}=64$) over the
complete binary slot-composition image and the learned subpixel position embedding, with a center-prior weight of $5.0$,
and the cross-attention proposal uses a 54-dimensional hidden MLP and a
32-dimensional query/key space.

\paragraph{Feature-similarity-masked equivariant convolutions.}
Each convolution forms local context only from neighbors whose latent direction is aligned with
the center pixel, using a cosine-similarity threshold $\tau_{\mathrm{cos}}=0.97$ between the
locally isometric feature vectors. This cosine gate is used as a
scale-invariant feature-direction mask; the calibrated angular neighborhoods are the L2
thresholds in equation~\ref{eq:ocrp_cluster_l2}. The masked neighborhood average is
combined with the center feature through an e3nn fully connected tensor
product~\cite{geiger2022e3nn} followed by a residual connection, keeping the update exactly
equivariant while suppressing crystallographically dissimilar neighbors. The IN718
configuration uses one $5\times5$ LR convolution with residual weight $1.0$, followed by two
$7\times7$ HR refinements with residual weight $0.3$; the \TiDataset{} configuration uses
one $3\times3$ LR convolution with residual
weight $1.0$, followed by one $3\times3$ HR refinement with residual weight $0.2$, avoiding
broader post-routing smoothing on the finer HCP boundary network.

\paragraph{Local support banks and feature-space clustering.}
For each LR pixel $p$, SG-SRAN extracts co-registered clustering and value banks
\begin{equation}
  \begin{aligned}
  &\\[-1pt]
  &\mathcal{B}^{(0)}_p = \{Z_{0,p,m}\}_{m=1}^{W_s^2},\quad
  \mathcal{B}^{(1)}_p = \{Z_{1,p,m}\}_{m=1}^{W_s^2},\\[-1pt]
  &W_s \in \{5,9\}\ \text{in the reported configurations},
  \end{aligned}
  \label{eq:ocrp_bank}
\end{equation}
with replicate padding at boundaries. The frozen encoded features $Z_0$ determine cluster membership,
whereas the context-refined equivariant features $Z_1$ provide the values later synthesized into HR tokens.
At $4\times$ upsampling this LR window spans roughly
$4W_s$ HR pixels per side (for example, $36$ HR pixels when $W_s=9$ and $20$ when $W_s=5$).
The $W_s^2$ nodes form an 8-neighbor graph in which two
adjacent nodes are linked when
\begin{equation}
  \|Z_{0,p,m} - Z_{0,p,m'}\|_2 \le \tau_c, \qquad
  \tau_c \in \{0.0349,\,0.0873\},
  \label{eq:ocrp_cluster_l2}
\end{equation}
corresponding under the local isometry to $2^\circ$ in the IN718 configuration and $5^\circ$ in
the \TiDataset{} configuration. Connected components of this
graph define local orientation regions; the largest $K$ components are packed into deterministic
slots, and for each slot $k$ SG-SRAN stores a binary membership mask $M_{p,k,m}$ and a compact
metadata vector $\mu_{p,k}$ (slot validity, slot-rank one-hot code, relative mass, centroid and
spatial dispersion).

\paragraph{Token-conditioned cross-attention proposal.}
SG-SRAN synthesises each candidate HR-token feature using a single token-conditioned
cross-attention operation over the members of a local orientation slot. Let the emitted patch be
indexed by $t \in \{1,\dots,T\}$, with $T=16$ for the $4\times4$ task, and let $\phi_t \in \RR^{32}$ denote the learned subpixel position embedding for token $t$. For slot $k$ at
LR site $p$, define its member set as
$\mathcal{S}_{p,k}=\{m : M_{p,k,m}=1\}$. A query constructed from the slot metadata
$\mu_{p,k}$ and the learned subpixel position embedding $\phi_t$ attends over these members; keys are derived from member
window coordinates and invariant norms of the context-refined member features, while values are the
context-refined member features $Z_1$:
\begin{equation}
  \widetilde{z}_{p,k,t}
  = \sum_{m \in \mathcal{S}_{p,k}} \beta_{p,k,t,m}\, Z_{1,p,m},
  \label{eq:ocrp_pool}
\end{equation}
where $\beta_{p,k,t,m}$ is a masked softmax over $m \in \mathcal{S}_{p,k}$. This directly produces an anchorless, token-conditioned proposal for each slot.
Since the attention
weights are invariant scalars and the values are equivariant features, $\widetilde{z}_{p,k,t}$ is
equivariant. This single attention operation performs both pooling and proposal generation,
allowing different tokens within the same HR patch to attend to different members of the same
local orientation cluster.

\paragraph{Geometric routing and owner selection.}
For each token the router emits logits over slots from the complete binary slot-composition image
$M_p\in\{0,1\}^{K\times W_s^2}$ and the token position,
\begin{equation}
  \begin{aligned}
  &\\[-1pt]
  &
  \boldsymbol{\ell}_{p,t}
  = f_{\mathrm{route}}\!\bigl(\mathrm{vec}(M_p),\, \phi_t\bigr)
  + w_c\,\mathbf{c}_p,
  \end{aligned}
  \label{eq:ocrp_router}
\end{equation}
where $\phi_t$ is the
learned subpixel position embedding for token $t$, $c_{p,k}=1$ when slot $k$ contains the
window center and is zero otherwise, and $w_c = 5.0$ biases the decision toward the center slot when occupancy evidence is
weak. The reported MLP router jointly processes all $K W_s^2$ mask entries and the token-position embedding, making routing geometric and separate from orientation-feature proposal synthesis; logits for empty slots are masked before selection. A hard owner $o_{p,t} = \arg\max_k \ell_{p,t,k}$ selects the slot for token $t$, implemented
with a straight-through estimator so gradients flow through the logits, and the emitted token is
the cross-attention proposal of that slot,
\begin{equation}
  y_{p,t} = \widetilde{z}_{p,o_{p,t},t}.
  \label{eq:ocrp_select}
\end{equation}
The $T$ selected tokens are reshaped into the HR patch for site $p$ and used directly as the HR
latent prediction; residual connections appear only within the downstream HR refinement
convolutions, after routed token selection.

\subsection{Dictionary-based decoder and fundamental-zone canonicalization}

The dictionary-based decoder maps a predicted latent feature $z \in \RR^C$ back to a unit quaternion in two steps.
First, a precomputed lookup table of $T=857{,}973$ fundamental-zone quaternions sampled by
cubochoric sampling~\cite{rosca2014cubochoric,rowenhorst2015msmse} is encoded once, $f_t = E(q_t)$, and the $k=2$
nearest table entries to $z$ are retrieved by chunked squared-Euclidean search. Second, starting
from these seeds, quaternion candidates are refined in parallel by Adam ($60$ steps, learning rate
$0.03$) to minimize $\|E(q) - z\|_2^2$, with $q$ renormalized to the unit sphere after each step,
and the candidate with the lowest residual is selected and canonicalized to the fundamental zone
by equation~\ref{eq:fz_reduce}. The finite cubochoric dictionary sets the small frozen-interface floor reported in Table~\ref{tab:main_results}a; that floor is below the downstream SR errors and is separated from the trainable upsampler when reporting representation fidelity. %
The encoder and dictionary-based decoder are frozen throughout training; only the
SG-SRAN upsampler and the refinement convolutions are optimized. The standalone representation-fidelity
result in Table~\ref{tab:main_results}a reports the dense lookup-only floor, isolating the frozen
encoder--dictionary-based decoder interface.

On an NVIDIA A100-PCIE-40GB, end-to-end IN718 $4\times4$ inference takes
$4.65$~s per $256\times256$ HR patch. The dictionary-based decoder takes ${\sim}4.63$~s
($70.6\,\mu\mathrm{s}$ per HR pixel), compared with $0.70$~ms for the encoder and $27.8$~ms for
the SR backbone. The dictionary search and refinement therefore set the current inference latency.

\subsection{Training objective and implementation}

The SG-SRAN backbone is trained with a feature-space mean-squared-error loss,
\begin{equation}
  \mathcal{L}_{\mathrm{feat}} = \frac{1}{B N_\mathrm{HR}} \sum_{b,n}
  \bigl\|\widehat{Z}_{\mathrm{HR},b,n} - E(Q_\mathrm{HR})_{b,n}\bigr\|_2^2,
  \label{eq:loss}
\end{equation}
where $\widehat{Z}_\mathrm{HR} = \mathcal{F}_{\mathrm{SG\mbox{-}SRAN}}(E(Q_\mathrm{LR}))$ is the predicted HR
latent from the full pipeline (equation~\ref{eq:ocrp_pipeline}), $E(Q_\mathrm{HR})$ is the frozen
encoded HR target (computed under \texttt{torch.no\_grad()} and detached), and $N_\mathrm{HR}$ is
the number of HR pixels per sample. By equation~\ref{eq:local_iso}, minimizing
$\mathcal{L}_{\mathrm{feat}}$ is a first-order surrogate for the squared geodesic orientation error,
\begin{equation}
  \mathcal{L}_{\mathrm{feat}} \;\approx\;
  \frac{1}{BN_\mathrm{HR}}\sum_{b,n}
  d_\Stab\!\bigl(\hat{q}_{b,n},\,q_{\mathrm{HR},b,n}\bigr)^2
  + O(\theta^3),
  \label{eq:loss_geo}
\end{equation}
with $\theta$ the per-pixel misorientation angle; the correspondence is exact to first order in
the locally isometric small-angle regime and remains an effective objective empirically at larger angles.
The backbone is trained with this loss alone, and the geometric router and its straight-through
owner assignment are learned end to end from the feature-matching gradient.

The model is implemented in PyTorch with e3nn~\cite{geiger2022e3nn} for the equivariant tensor
products. At run time, Reynolds invariant features are evaluated from precomputed projector
bases using spherical-harmonic synthesis of $D^l(R(q))U_l^{(\Stab)}$, entirely on GPU; the
Wigner-$D$ eigensolve and angular-momentum-generator machinery enters only in the offline
construction of the projectors and bases. Optimization uses AdamW (weight decay
$10^{-4}$) at peak learning rate $3\times10^{-4}$ with cosine annealing ($2$ warmup epochs, minimum
learning rate $10^{-6}$), gradient clipping at norm $1.0$, and bfloat16 mixed precision with
TensorFloat-32, for $150$ epochs at batch size $2$ on IN718 and $5$ on \TiDataset{}. The aggregate benchmark reports five
independent SG-SRAN seeds ($42$--$46$); seed $42$ is used for the representative qualitative
figures and mechanism analysis.

\subsection{Datasets and evaluation protocol}

\paragraph{Materials.}
Two EBSD datasets are used for in-domain training and testing. The IN718 source map is an
EBSD orientation field from the multi-modal 3D microstructure dataset of Stinville et al.~\cite{stinville2022multimodal};
the \TiDataset{} source data are drawn from the Ti-6Al-4V 3D microstructure dataset of Jangid et al.~\cite{jangid2023titanium3d}. The IN718 dataset (nickel superalloy, FCC,
$\Stab = O$) provides $1174/147/147$ train/validation/test paired patches at HR size
$256 \times 256$; the \TiDataset{} dataset (titanium alloy, HCP, $\Stab = D_6$) provides
$581/73/72$ train/validation/test paired patches at HR size $128 \times 128$. SG-SRAN is
trained and evaluated end to end on both materials under the $r=(4,4)$ task. The frozen
encoder--dictionary-based decoder round-trip baseline in Table~\ref{tab:main_results}a is reported on the
IN718 and \TiDataset{} test splits as an FCC/HCP representation analysis rather than as a trained SR result. The dataset manifests record
patch counts, sizes, and crystal symmetry; train, validation, and test splits are disjoint at
the patch-file level. The zero-shot transfer table uses two held-out targets: CoNi (FCC) and
Ti-Al (HCP). The CoNi scan is a held-out FCC transfer map prepared with the same manifest
workflow, and the Ti-Al target uses an HCP Ti-Al split with $1064/133/133$
train/validation/test paired patches; only its test split is used for zero-shot evaluation.

\paragraph{Training-pair construction.}
For each source map, quaternions are normalized, mapped to a consistent hemisphere, and reduced to
the fundamental zone. Random aligned HR patches are extracted, and the matching LR patches are
formed by regular stride-$r$ subsampling along each axis ($r=4$), so LR and HR remain spatially
registered without an extra interpolation step. This construction models the spatial-resolution
component of the LR/HR trade-off.

\paragraph{Baselines.}
We compare against four classical interpolants. \emph{Nearest}:
nearest-neighbor upsampling with quaternion renormalization. \emph{Bicubic}: per-channel bicubic
interpolation, renormalized to unit norm. \emph{SLERP}: pairwise spherical linear
interpolation~\cite{shoemake1985slerp} on the quaternion sphere. \emph{Symmetry-aware SLERP}:
bilinear SLERP in which each pair is first symmetry-resolved to the shortest crystallographic path.
All four are generated from the same LR test patches used by SG-SRAN. Seven learned
baselines are trained independently on the matched IN718 and \TiDataset{}
train/validation splits and evaluated on the corresponding $n=147$ and $n=72$
held-out test patches at $4\times4$, each over five training seeds (42--46).
\qrbsaadapted~\cite{jangid2024qrbsa} retains the released
10-block, 128-feature quaternion residual self-attention head and body. Its local adapter
converts the passive scalar-first dataset quaternions to the active scalar-last convention
expected by the released implementation, selects the 24-element proper $O$ subgroup
for IN718 or the 12-element proper $D_6$ subgroup for \TiDataset{} from the dataset
point-group metadata for the rotational loss, and exports predictions through the
same fundamental-zone and held-out evaluation protocol as SG-SRAN. The $4\times4$ model replaces the released one-dimensional pixel-shuffle tail
with a dimension-matched two-dimensional quaternion pixel-shuffle tail. It is trained from scratch for 300 epochs and contains $6.17$ million parameters, using AdamW with cosine decay (peak learning rate $3\times10^{-4}$, weight
decay $10^{-4}$, two warmup epochs, seeds 42--46).

QEDSR is a quaternion adaptation of EDSR~\cite{lim2017edsr}: the standard EDSR
residual body operates on four quaternion channels (active scalar-last, with the passive
scalar-first convention restored after inference), followed by a two-dimensional quaternion
pixel-shuffle tail. It contains $1.82$M trainable parameters and is trained from scratch for
$150$ epochs with AdamW (peak learning rate $3\times10^{-4}$, weight decay $10^{-4}$, two
warmup epochs, five seeds 42--46) under the same dataset-selected $O$ or $D_6$ rotational
loss and held-out evaluator as SG-SRAN.
The corresponding real-valued EDSR baseline uses the same residual-block schedule and
two-dimensional pixel-shuffle tail on four scalar quaternion channels, without quaternion
convolutions or quaternion attention. It contains $7.24$M trainable
parameters and is trained and evaluated with the same five-seed held-out protocol as the
other learned baselines.

RCAN~\cite{zhang2018rcan}, SAN~\cite{dai2019san} and
HAN~\cite{niu2020han} retain their published residual-channel, second-order and holistic
attention bodies, respectively. RGB mean-shift layers are omitted and active scalar-last
quaternions are supplied as four channels; the passive scalar-first convention is restored
after inference. RCAN uses 10 residual groups with 20 blocks each, SAN uses 20 groups with
10 blocks each, and HAN uses 10 groups with 20 blocks each, all with 64 features. The $4\times4$ models use a two-dimensional pixel-shuffle tail; parameter counts are $15.59$M (RCAN), $15.86$M (SAN) and $16.07$M (HAN).
Each dataset is trained independently for 300 epochs with AdamW, cosine decay,
weight decay $10^{-4}$, two warmup epochs and five seeds (42--46). SAN uses its published
$10^{-4}$ base rate on both materials; RCAN/HAN use $3\times10^{-4}$ on IN718 and the
authors' $10^{-4}$ rate on \TiDataset{} after the larger rate proved unstable under the
proper-$D_6$ loss. SAN covariance pooling is evaluated by the exactly equivalent centered
product $XX^\top/(HW)$ rather than materializing an $(HW)^2$ centering matrix, and its
trace-normalized Newton--Schulz square root uses native autograd with a
machine-precision diagonal regularizer. All three models use the same dataset-selected
$O$ or $D_6$ rotational loss and held-out evaluator as SG-SRAN.

The Atindama baseline adapts the partial-convolution U-Net from the EBSD
restoration pipeline of Atindama et al.~\cite{atindama2023restoration}. HR orientations are
converted to normalized intrinsic ZXZ Euler channels, the observed stride-$r$ samples define
a periodic known-pixel mask, and the network predicts the missing pixels while known samples
are copied exactly into the output. The $4\times4$ model is trained
for 150 epochs with Adam (learning rate $2\times10^{-4}$, five seeds 42--46) using squared $\SO$
geodesic error on unknown pixels without crystal-symmetry minimization. Training is
therefore symmetry-agnostic, but all exported Ti predictions are reduced and evaluated
with the same 12 proper $D_6$ rotations used by the other methods. Each model has
$25.78$ million trainable parameters. The published Criminisi refinement is not applied:
the regular SR masks contain no fully known $3\times3$ source patch, so its exemplar-search
precondition is unsatisfied.

An architecture-matched no-routing control isolates the routing mechanism: the encoder, refinement convolutions and dictionary-based decoder are fixed, and only the routed-patch upsampler is replaced by bicubic or nearest interpolation in the latent space. The control was run for the IN718 Reynolds $4\times4$ checkpoint under a single-seed protocol; its results are reported alongside the boundary
metrics in Results.

\paragraph{Evaluation metrics.}
Angular accuracy is measured by the symmetry-aware misorientation $d_\Stab$
(equation~\ref{eq:sym_err}). The FCC metric minimizes over the 24 proper rotations
of $O$, and the HCP metric minimizes over the 12 proper rotations of $D_6$, so each
material is evaluated with its own rotational crystal symmetry. For the frozen
encoder--dictionary-based decoder round trip, we report the mean and standard
deviation of the per-sample mean $d_\Stab$ in radians on the held-out IN718 and
\TiDataset{} test splits. For the super-resolution benchmark,
angular errors are reported in degrees and pooled over all HR pixels in the 147
held-out IN718 patches or the 72 held-out \TiDataset{} patches. The mean, median,
p68, p95 and p99 summarize this pooled per-pixel distribution; per-patch or
seed-level standard deviations are reported where uncertainty is shown.

To evaluate boundary recovery, we also split $d_\Stab$ into grain-interior and
grain-boundary-band pixels. The HR boundary mask is obtained by thresholding
4-connected nearest-neighbor crystallographic misorientation at $5^\circ$, and the
boundary band contains pixels within five pixels of this HR mask. Boundary F1 is
computed from predicted and HR boundary masks using the same $5^\circ$ rule, with
TP, FP and FN accumulated over all patches:
$F1 = 2\,\mathrm{TP}/(2\,\mathrm{TP}+\mathrm{FP}+\mathrm{FN})$.

Inverse-pole-figure (IPF) maps color each pixel by the crystal direction parallel
to a selected sample axis ($X$, $Y$ or $Z$), making grains and sharp orientation
changes visually legible; we render them with \textit{orix}~\cite{johnstone2020orix,orix0130}.
IPF-space PSNR and SSIM are computed on the IPF-$X$, $Y$ and $Z$ renderings of the
prediction against the HR reference and averaged over the three axes. All angular
metrics are computed directly on the quaternion orientation field.

For the boundary-window composition analysis, an 8-neighbor $5^\circ$
symmetry-aware rule defines HR boundary centers. Each center contributes clipped
$3\times3$ HR and SR windows. Transitive pairwise matches within $5^\circ$ partition
the HR window into orientation groups. The spurious rate is the fraction of SR
observations farther than $5^\circ$ from every HR group; composition recall is the
fraction of HR groups recovered by at least one SR observation; and Composition F1
is the harmonic mean of recall and precision, with precision $=1-$spurious rate.
Overlapping boundary windows are counted separately.

\backmatter

\bmhead{Data availability}
EBSD datasets and encoded lookup tables will be deposited in a public
repository upon publication. Requests for data prior to publication
should be directed to the corresponding author.

\bmhead{Code availability}
Code implementing the SG-SRAN-based latent super-resolution model, the
locally isometric encoder, and the dictionary-based decoder will be released
under the Apache~2.0 license on a public GitHub repository upon publication. The
release will include trained checkpoints for the IN718 and \TiDataset{}
$4\times4$ models, the precomputed Reynolds basis files,
and the cubochoric-sampled dictionary lookup tables ($T=857{,}973$ per symmetry class for FCC and HCP), which would
otherwise be non-trivial to recompute.

\bmhead{Acknowledgements}
The authors acknowledge Neal Brodnik (Postdoc, UCSB) for his intellectual contributions and early discussions on this problem. We also thank Joaquin Giorgio (M.S., UCSB) for his prior work
on SLERP-based orientation upsampling, which informed the baseline
comparison in this study. This research was supported in part by the NSF CSSI Award \#2411453. TMP and MPE also acknowledge the support of Army Research Office (ARO) MURI Grant W911NF-25-2-0164.

\section*{Competing interests}
The authors declare no competing interests.

\section*{Author contributions}

U.G. developed the direct Reynolds-projection embedding used in the final HCP and FCC representations, along with its calibration, and developed the anchorless routing architecture. U.G. led the experimental validation and conducted the experiments, and wrote the manuscript. W.Z. proposed applying the Reynolds operator to enforce symmetry-compliant network outputs and developed the original locally isometric embedding and its calibration, using rank-tensor decomposition. U.G. and W.Z. jointly developed and tested early implementations of this approach. Both authors revised the manuscript; U.G. verified the manuscript's claims against the implementation. M.P.E. acquired and provided EBSD datasets, contributed to experimental design, and reviewed the manuscript. S.H.D. provided guidance on data interpretation, and reviewed the manuscript. T.M.P. provided guidance on materials characterization and EBSD methodology, and reviewed the manuscript. B.S.M. supervised the overall project, provided guidance on the manuscript, and reviewed and edited the manuscript.

\bibliography{sn-bibliography}

\clearpage
\setcounter{figure}{0}
\renewcommand{\figurename}{Extended Data Fig.}
\renewcommand{\theHfigure}{ED\arabic{figure}}
\setcounter{table}{0}
\renewcommand{\tablename}{Extended Data Table}
\renewcommand{\theHtable}{EDT\arabic{table}}
\section*{Extended Data}
\makeatletter
\setlength{\@fptop}{0pt}
\setlength{\@fpbot}{0pt plus 1fil}
\makeatother

\begin{figure}[p]
\centering
\begin{overpic}[width=\textwidth]{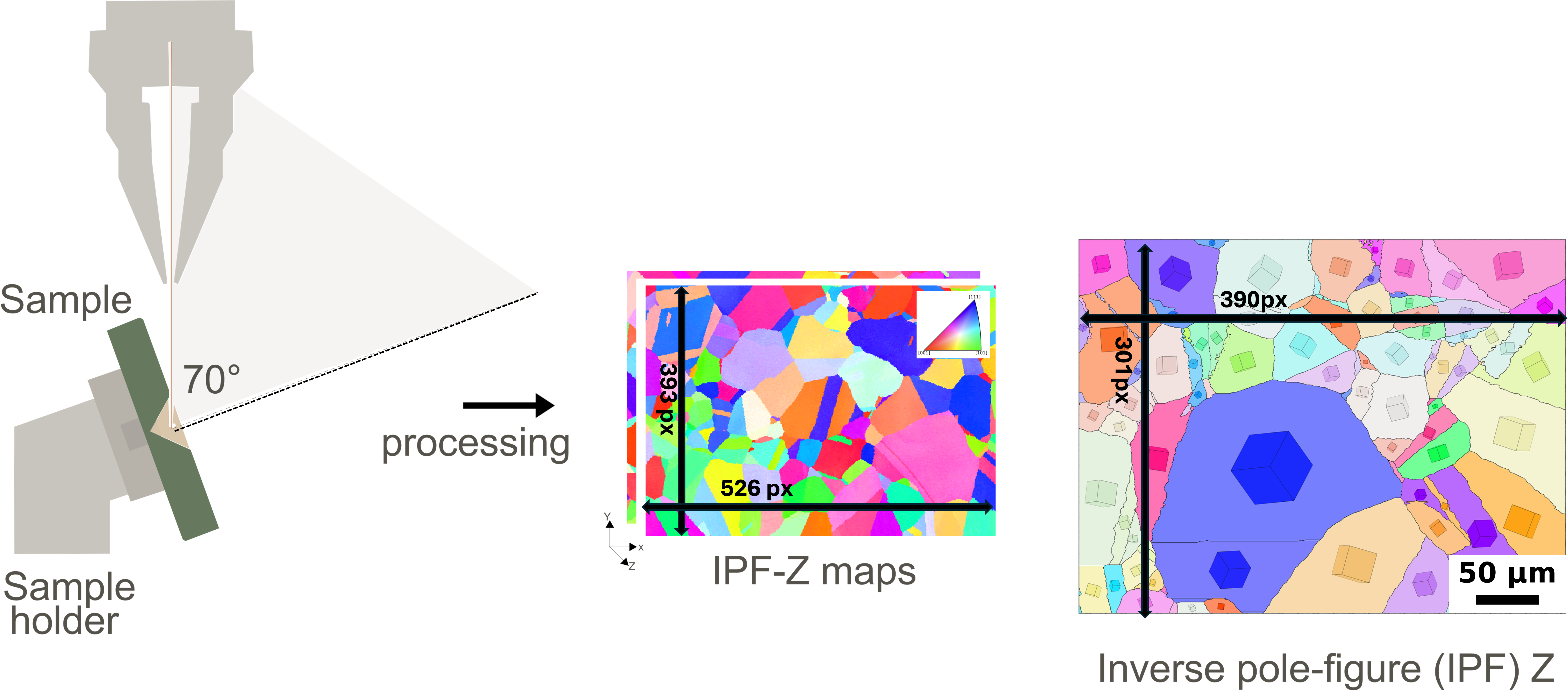}
  \put(0.5,40){\fontsize{10.5}{12.0}\selectfont\sffamily\bfseries a}
  \put(64,40){\fontsize{10.5}{12.0}\selectfont\sffamily\bfseries b}
\end{overpic}
\caption{\textbf{Electron backscatter diffraction (EBSD) and the inverse pole figure (IPF) representation.} \textbf{a}, A focused electron beam is scanned across a specimen tilted to $70^\circ$ in a scanning electron microscope, and the backscatter diffraction pattern collected at each scan point is indexed to a crystallographic orientation; the resulting orientation field is rendered as an inverse pole-figure (IPF-$Z$) map. \textbf{b}, In the IPF representation, color encodes the crystal direction aligned with the sample $Z$ axis according to the IPF color key (colored fundamental triangle). Grains appear as contiguous regions of slowly varying color separated by sharp orientation boundaries; recovering these boundaries at high spatial resolution, under a fixed acquisition budget, is the objective of this work.}\label{fig:ebsd_pipeline}
\end{figure}

\begin{figure*}[!t]
\centering
\includegraphics[width=\textwidth]{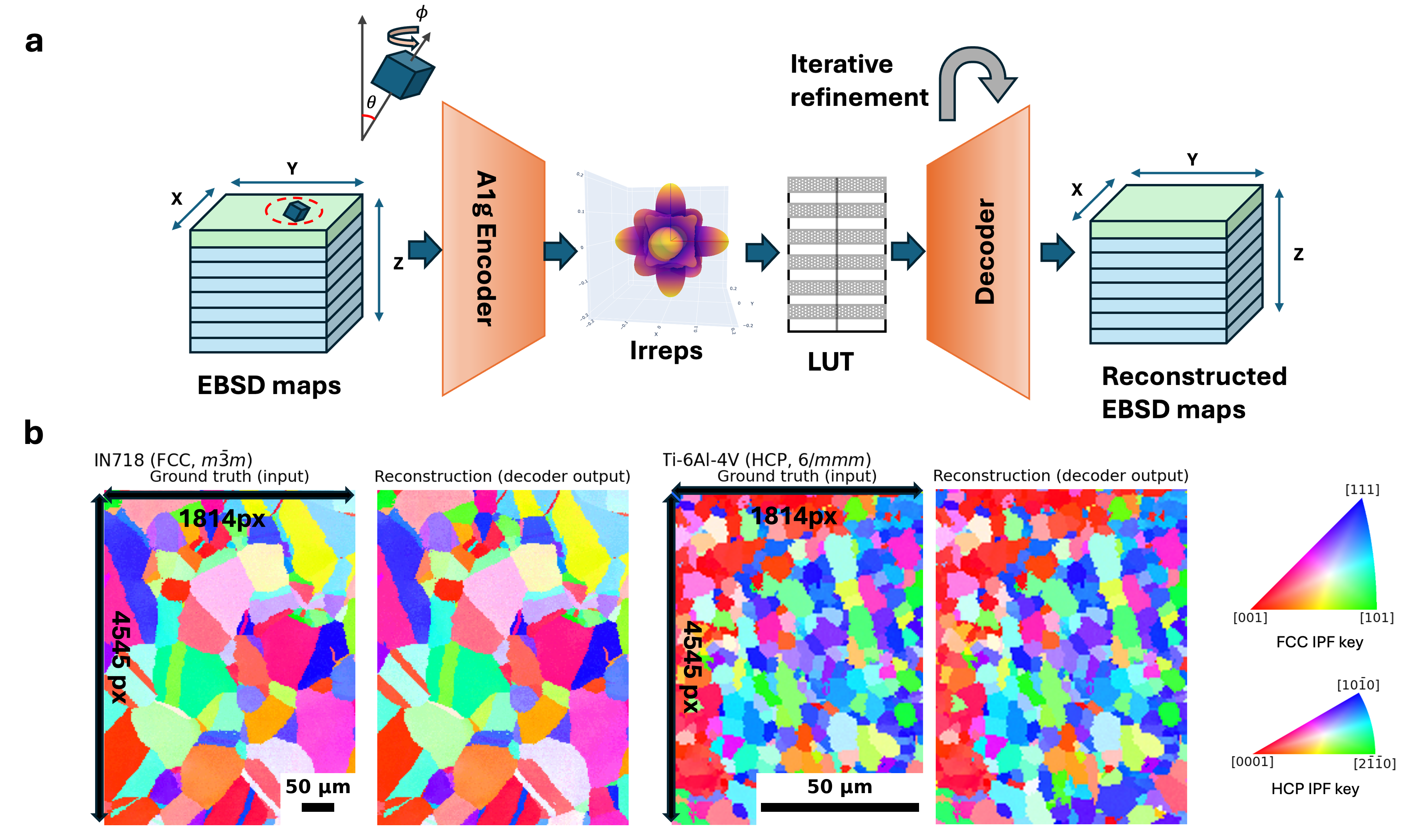}
\caption{%
\textbf{Frozen orientation interface and encoder--dictionary-based decoder reconstruction fidelity.}
\textbf{a}, Each pixel orientation is mapped by the Reynolds-projected $A_1$ encoder to
crystal-symmetry-invariant latent features. Predicted HR features are matched against
the dictionary, generated by cubochoric sampling, and decoded to unit quaternions, with optional local iterative
refinement. The $4\times4$ annotation denotes the surrounding SR task: spatial
upsampling is performed by the SG-SRAN backbone (Fig.~\ref{fig:arch}), whereas the frozen encoder and
dictionary-based decoder act pointwise on the LR and HR orientation fields.
\textbf{b}, Round-trip encode$\to$decode reconstructions on full IPF-$Z$ scan slices for FCC IN718
using the proper rotational subgroup $O$ (left pair) and \TiDataset{} HCP test data using
the proper rotational subgroup $D_6$ (right pair). In each pair, the
ground-truth input and dictionary-based decoder output are visually indistinguishable, including at grain
boundaries. The dense-lookup round-trip misorientation is approximately $0.45^\circ$
($d_\Stab$) for both systems (Table~\ref{tab:main_results}a), showing that the frozen latent
interface round-trips orientation information across the two symmetry classes. Panel b isolates
pointwise encoder--dictionary-based decoder reconstruction at unchanged spatial sampling.%
}\label{fig:reconstruction}
\end{figure*}

\begin{figure*}[p]
\centering
\begin{minipage}{\textwidth}\centering
{\sffamily\bfseries\makebox[\textwidth][l]{\fontsize{10.5}{12.0}\selectfont a\hspace{0.65em}\normalsize IN718 (FCC, $O$)}}\\[1pt]
\includegraphics[width=\textwidth]{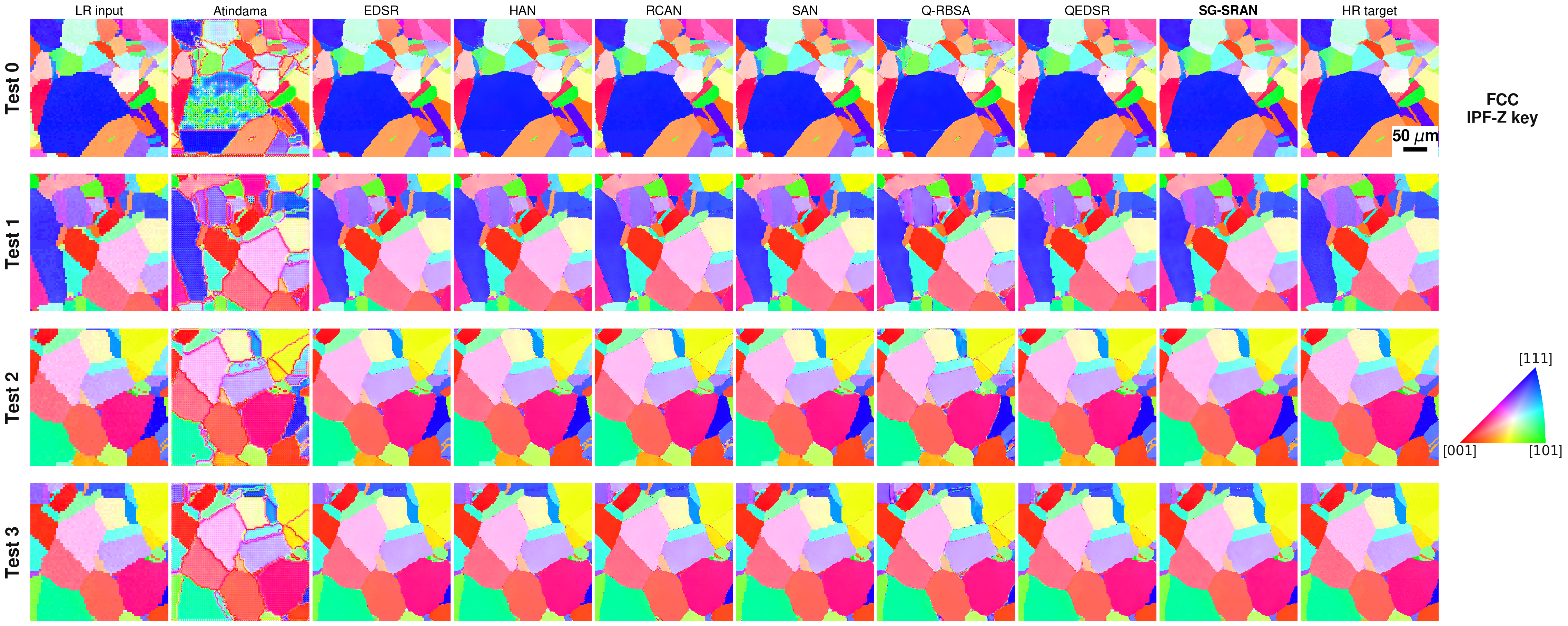}
\end{minipage}\\[4pt]
\begin{minipage}{\textwidth}\centering
{\sffamily\bfseries\makebox[\textwidth][l]{\fontsize{10.5}{12.0}\selectfont b\hspace{0.65em}\normalsize \TiDataset{} (HCP, $D_6$)}}\\[1pt]
\includegraphics[width=\textwidth]{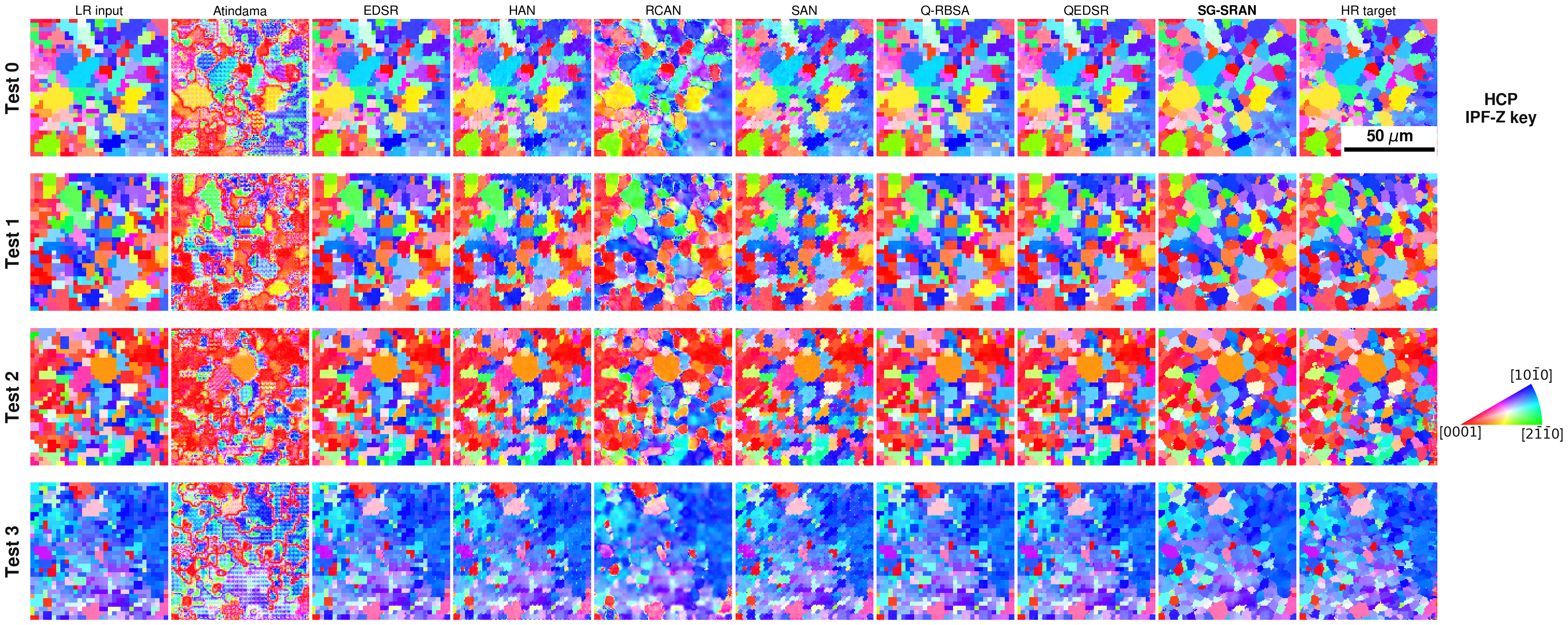}
\end{minipage}
\caption{\textbf{Expanded learned-baseline comparison on IN718 and \TiDataset{} held-out $4\times4$ examples.}
\textbf{a}, IN718 comparison with the full Open718 block~\cite{stinville2022multimodal} labeled Test 0, followed by
held-out test samples 1--3, shown at a common extent and aspect ratio. \textbf{b}, \TiDataset{}
trained-baseline comparison on held-out test samples 0--3 from the HCP split.
In both panels, columns show the display-upsampled LR input, Atindama inpainting,
EDSR, HAN, RCAN, SAN, \qrbsaadapted, QEDSR, SG-SRAN and the HR target, with the
material-specific IPF-$Z$ key at right. One displayed LR pixel
corresponds to a $4\times4$ HR block.}\label{fig:new_learned_baselines}
\end{figure*}

\begin{figure}[p]
\centering
\begin{minipage}{\textwidth}\centering
{\sffamily\bfseries\makebox[\textwidth][l]{\fontsize{10.5}{12.0}\selectfont a\hspace{0.65em}\normalsize IN718 (FCC, $O$)}}\\[1pt]
\includegraphics[width=0.98\textwidth]{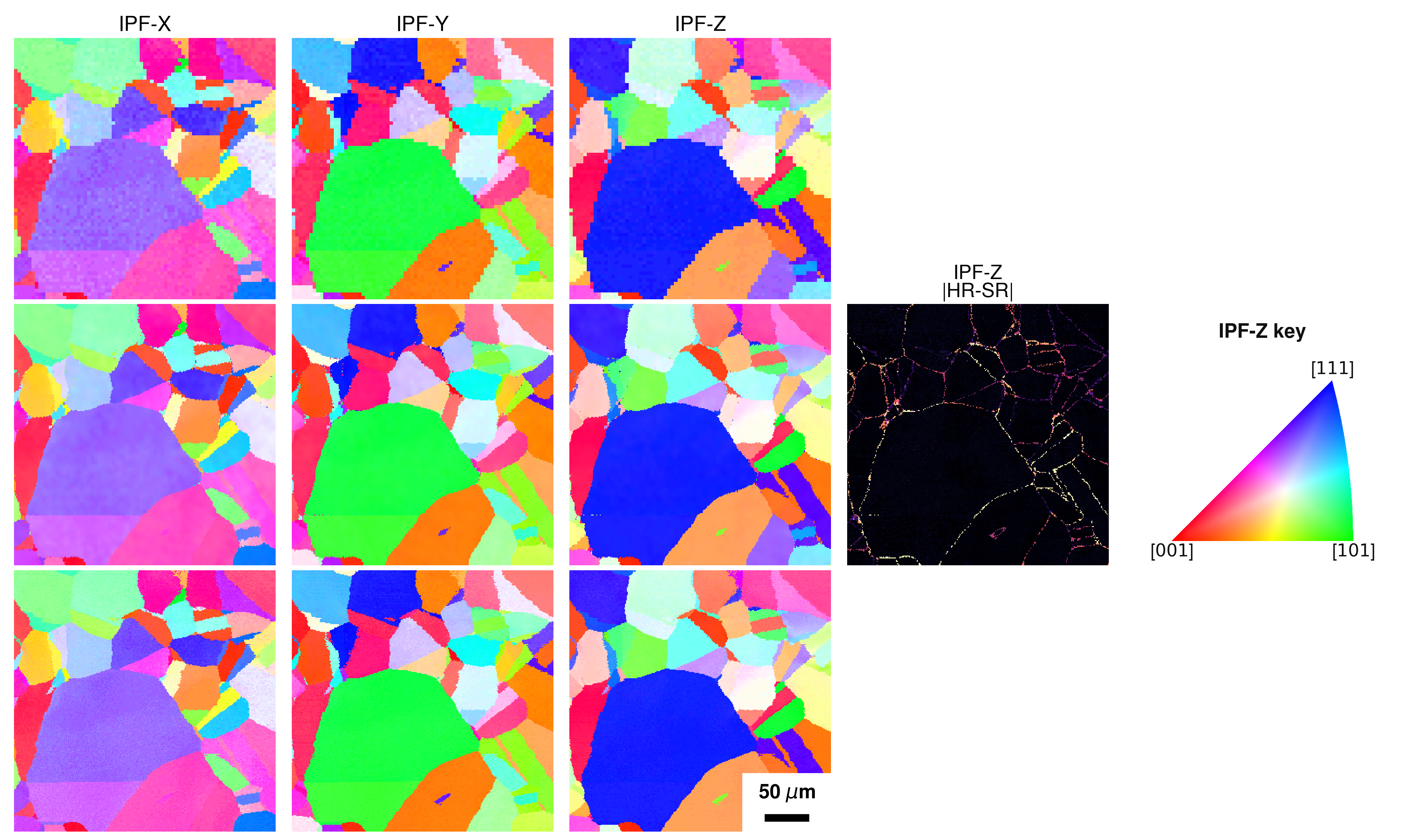}
\end{minipage}\\[3pt]
\begin{minipage}{\textwidth}\centering
{\sffamily\bfseries\makebox[\textwidth][l]{\fontsize{10.5}{12.0}\selectfont b\hspace{0.65em}\normalsize \TiDataset{} (HCP, $D_6$)}}\\[1pt]
\includegraphics[width=0.98\textwidth]{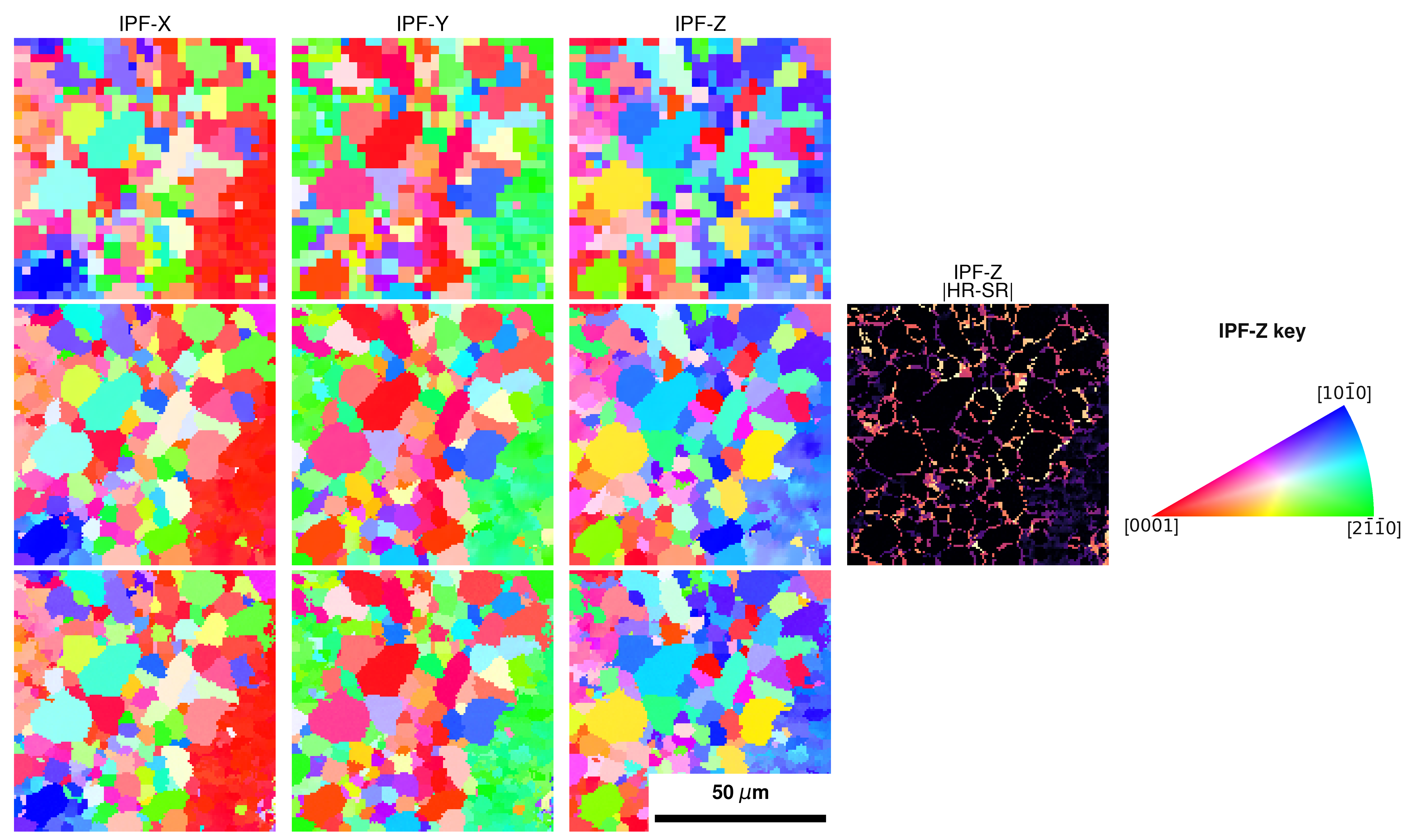}
\end{minipage}
\caption{%
\textbf{Held-out $4\times4$ test samples on both crystal systems.}
Within each material panel, from top to bottom, rows show the display-upsampled low-resolution input,
the SG-SRAN super-resolved output and the high-resolution target; the first three columns show
IPF-$X$, IPF-$Y$ and IPF-$Z$, and the fourth visual column shows the IPF-$Z$ residual
$|\mathrm{HR}-\mathrm{SR}|$ with brighter pixels indicating larger residual.
The LR input is displayed at the HR grid via nearest-neighbor upsampling only for
visual alignment; the SG-SRAN SR and HR fields are at the native HR resolution. (\textbf{a})
IN718 held-out test sample. (\textbf{b}) \TiDataset{} held-out test sample from the
HCP test split. One LR pixel corresponds to a $4\times4$ HR block; the displayed HR fields
are $256\times256$ pixels for IN718 and $128\times128$ pixels for \TiDataset{}.%
}\label{fig:visual}
\end{figure}

\begin{figure}[p]
\centering
\includegraphics[width=0.94\textwidth]{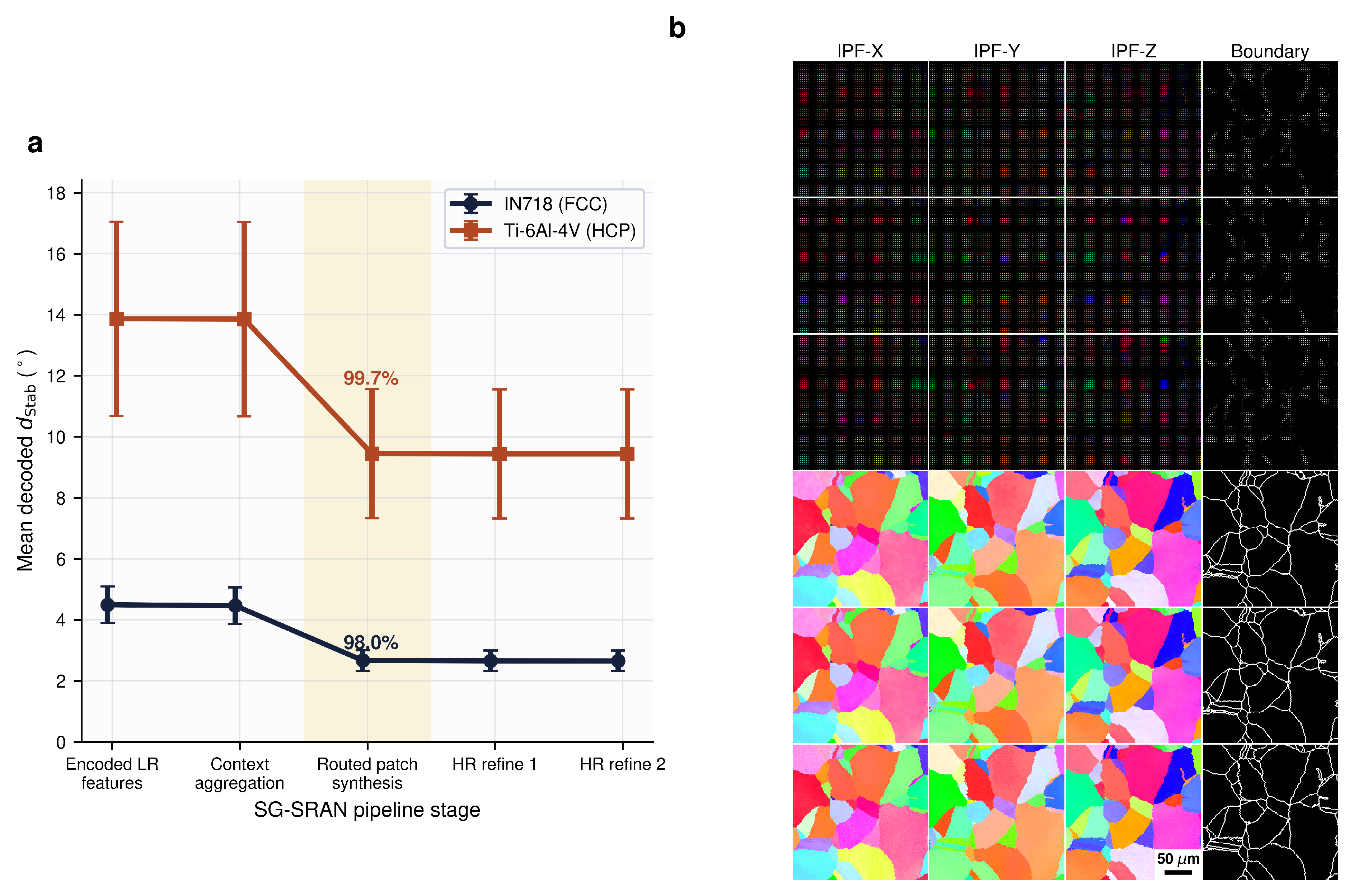}
\caption{%
\textbf{Stagewise decoded error and sample-level interpretability for SG-SRAN.}
\textbf{a}, Mean decoded symmetry-aware misorientation for intermediate latent features from
the seed-42 SG-SRAN checkpoints on IN718 and \TiDataset{}. The routed-upsampler decrease accounts
for $98.0\%$ of the total encoded-LR-to-final decrease on IN718 and
$99.7\%$ on \TiDataset{};
error bars show the standard deviation across per-sample means. \textbf{b}, IN718 Test
sample 0 decoded through the SG-SRAN pipeline. From top to bottom, the rows show the observed
LR samples, decoded LR embedding, context aggregation, routed patch synthesis, final SG-SRAN output and HR target.
Columns show IPF-$X$, IPF-$Y$, IPF-$Z$ and the $5^\circ$ boundary map. LR-space rows are
displayed on the HR canvas only at sampled sites; unsampled pixels are black. The IN718
test canvas is $256\times256$ HR pixels, with LR rows sampled every fourth HR pixel.%
}\label{fig:stagewise_progression}
\end{figure}

\begin{figure}[p]
\centering
\includegraphics[width=0.98\textwidth]{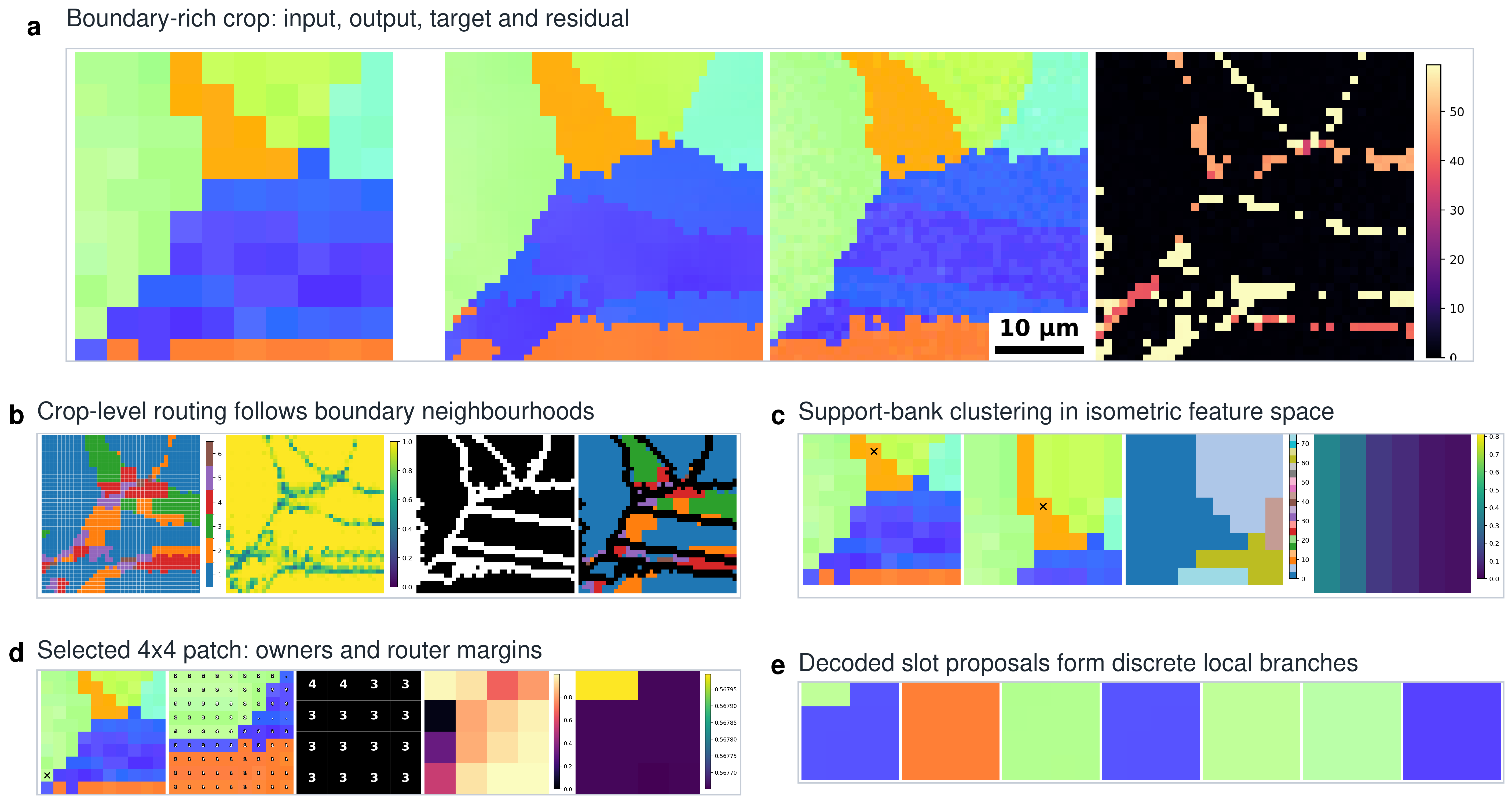}
\caption{%
\textbf{SG-SRAN local-routing walkthrough on a boundary-rich IN718 test crop.}
\textbf{a},
Boundary-rich crop context: the four tiles show the sparse LR input, the decoded SG-SRAN SR map,
the HR target and the SR--HR misorientation map, localizing residual error to grain-boundary
neighborhoods.
\textbf{b},
Crop-scale routing analysis: owner-slot assignments, router confidence, the HR boundary mask and
the owner-boundary overlay show that route changes and confidence drops align with grain-boundary
neighborhoods.
\textbf{c}, $9\times9$
Local support-bank decomposition for the marked LR site: the configured local window-support bank,
shown here as a $9\times9$ IN718 neighborhood, is grouped into orientation-consistent slots in the isometric latent feature space, and the
rightmost bars report the center-site slot masses. They summarize local support composition for interpretation; the router itself consumes the complete binary slot masks and token position.
\textbf{d}, $4\times4$
Patch-level routing decision for the selected $4\times4$ HR block: crop context,
per-pixel owner slots, selected owners and router margins show which support branch is selected at
each HR location.
\textbf{e},
Decoded candidate features: each candidate slot is mapped back to an IPF-colored orientation
branch, making residual errors interpretable as branch-selection errors among discrete local
candidate features rather than unconstrained quaternion averaging.
The displayed $9\times9$ LR support footprint spans $36\times36$ HR pixels at $4\times$ upsampling,
and the selected routed output block is $4\times4$ HR pixels.}%
\label{fig:probe_routing}
\end{figure}

\begin{figure}[p]
\centering
\includegraphics[width=\textwidth]{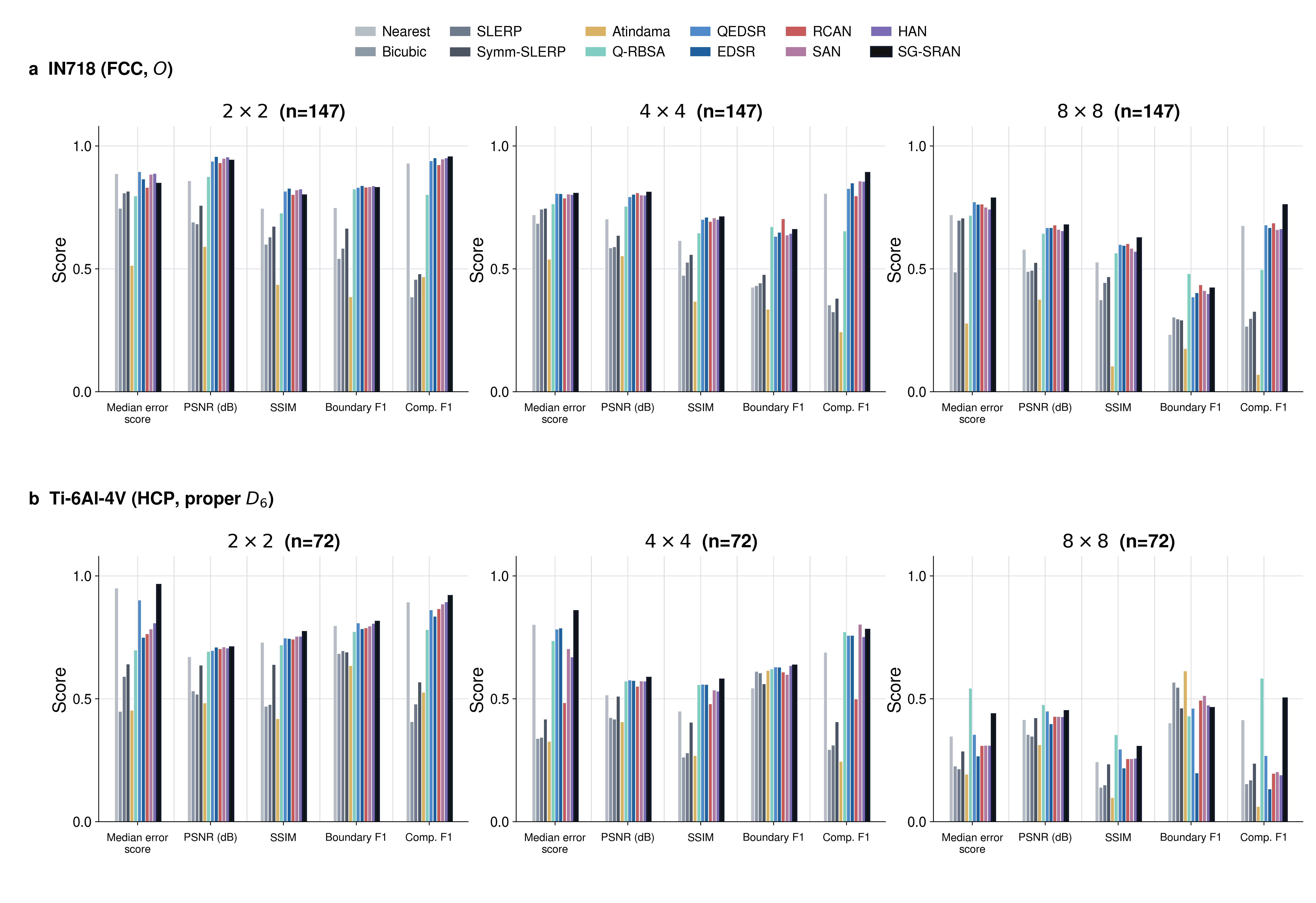}
\caption{%
\textbf{Scale dependence across crystal systems and upsampling factors.}
Seed-42 performance is shown for deterministic interpolants, learned baselines and SG-SRAN
as grouped normalized score bars at $2\times2$, $4\times4$ and $8\times8$. \textbf{a}, IN718
(FCC, $O$). \textbf{b}, \TiDataset{} (HCP, proper $D_6$). Each scale subplot reports median
symmetry-aware misorientation, IPF-space PSNR, IPF-space SSIM, Boundary F1 and
boundary-window Composition F1. Because these metrics have different units, values are
normalised to $[0,1]$ against fixed physical anchors rather than subplot-wise
min--max ranges, after orienting all metrics so higher bars indicate better
performance: median misorientation is inverted on a logarithmic scale between
$0.2^\circ$ and the crystal-symmetry covering radius ($62.8^\circ$ for cubic $O$ and
$93.8^\circ$ for hexagonal $D_6$), PSNR is linear on $[0,25]$~dB, and SSIM, Boundary F1
and Composition F1 are linear on $[0,1]$. Each bar is evaluated from the corresponding
scale-specific prediction summary; SG-SRAN uses the reported scale-specific invariant-embedding router configuration.%
}\label{fig:scale_dependence}
\end{figure}

\begin{figure*}[p]
\centering
\begin{minipage}{\textwidth}\centering
{\sffamily\bfseries\makebox[\textwidth][l]{\fontsize{10.5}{12.0}\selectfont a\hspace{0.65em}\normalsize CoNi zero-shot (FCC, $O$)}}\\[1pt]
\includegraphics[width=\textwidth]{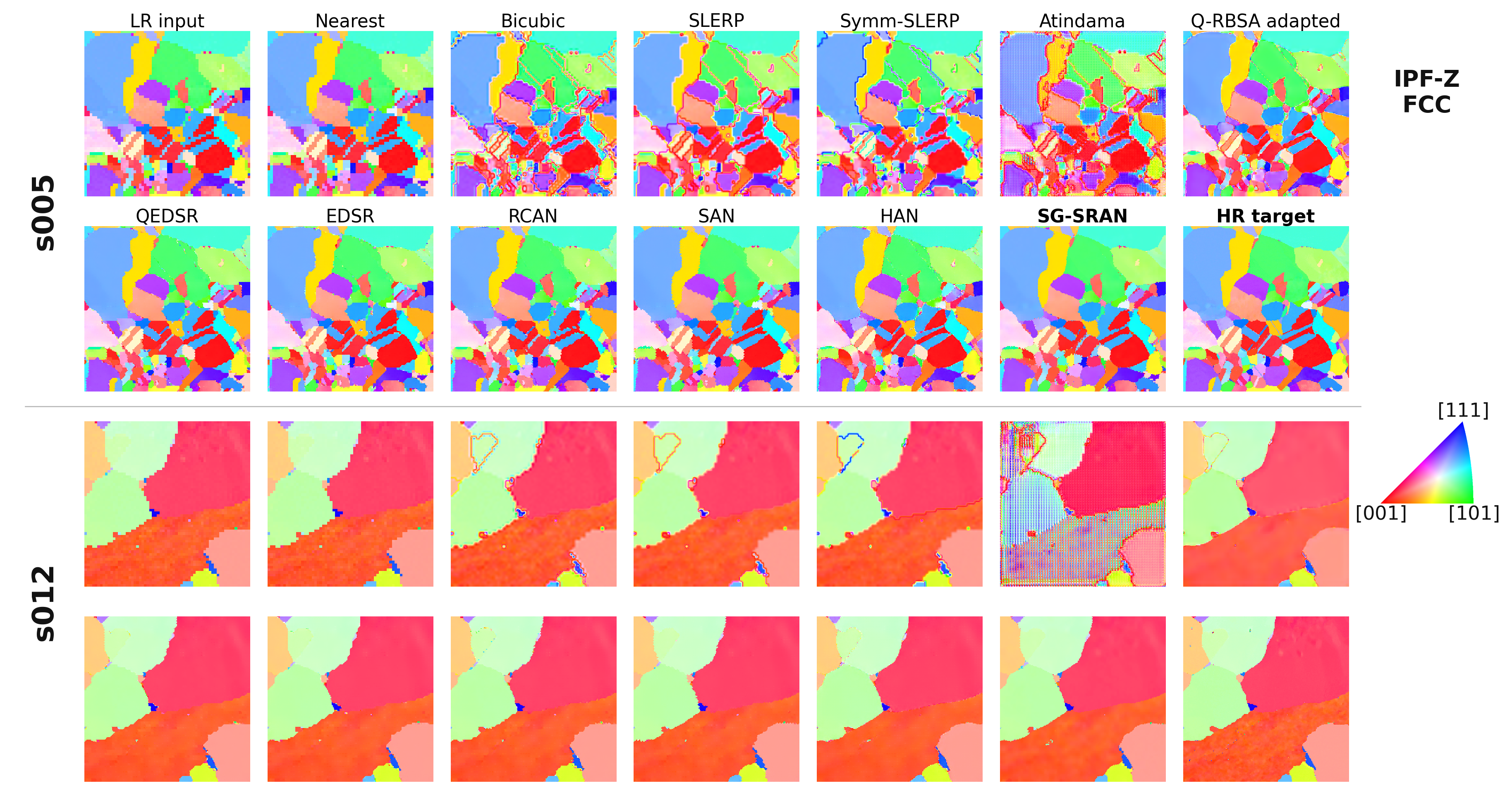}
\end{minipage}\\[5pt]
\begin{minipage}{\textwidth}\centering
{\sffamily\bfseries\makebox[\textwidth][l]{\fontsize{10.5}{12.0}\selectfont b\hspace{0.65em}\normalsize Ti-Al zero-shot (HCP, $D_6$)}}\\[1pt]
\includegraphics[width=\textwidth]{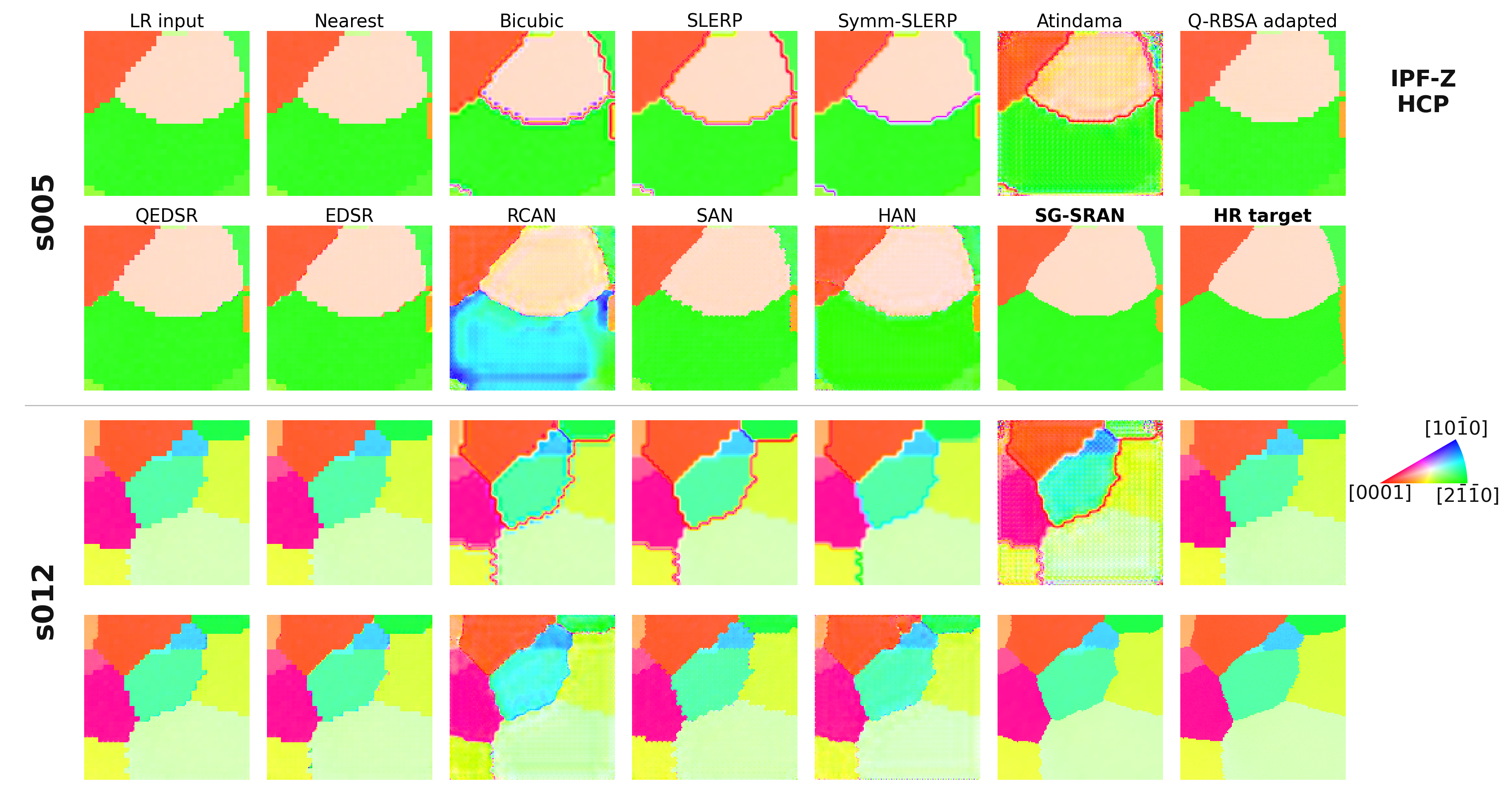}
\end{minipage}
\caption{\textbf{Zero-shot IPF-$Z$ outputs across all baselines for FCC and HCP transfer targets.}
\textbf{a}, IN718-trained models evaluated without target retraining on two CoNi samples.
\textbf{b}, Ti-6Al-4V-trained models evaluated without target retraining on two Ti-Al samples.
Within each material panel, two seven-column rows show the display-upsampled LR input, deterministic
interpolants, learned quaternion or image-space baselines, SG-SRAN and the HR target.
The LR input is nearest-neighbor expanded only for visual alignment; all SR and HR panels
are rendered at the native HR grid from saved quaternion arrays, with SLERP and
Symm-SLERP recomputed from the same LR samples at render time. Each subfigure has one material-appropriate Laue
IPF-$Z$ key, while the misorientation/evaluation protocol uses the rotational subgroups $O$ and
$D_6$.}\label{fig:zero_shot_ipfz_all_methods}
\end{figure*}

\clearpage

\begin{table*}[p]
\caption{\textbf{Boundary-window composition analysis on in-distribution test splits.}
This pooled diagnostic is evaluated on the selected prediction summaries used for the composition analysis. For each HR boundary pixel, the HR and SR $3\times3$ windows centered at that pixel are compared under crystal symmetry with a $5^\circ$ tolerance. The spurious rate is the fraction of SR window observations not matching any GT orientation group in the corresponding HR window (lower is better). Recall is the fraction of GT orientation groups recovered by at least one SR observation, and Composition F1 uses precision $=1-$spurious rate together with this recall. Boundary centers use the HR 8-neighbor $5^\circ$ boundary mask; overlapping boundary windows are counted separately.}\label{tab:boundary_composition_indist}
\centering
{\scriptsize
\setlength{\tabcolsep}{3.2pt}
\resizebox{\textwidth}{!}{%
\begin{tabular}{lcccccc}
\toprule
Method &
\multicolumn{3}{c}{IN718 $4\times4$} &
\multicolumn{3}{c}{\TiDataset{} $4\times4$} \\
\cmidrule(lr){2-4}\cmidrule(lr){5-7}
& Spurious $\downarrow$ & Recall $\uparrow$ & F1 $\uparrow$ &
Spurious $\downarrow$ & Recall $\uparrow$ & F1 $\uparrow$ \\
\midrule
Nearest & \textbf{0.030} & 0.691 & 0.807 & 0.149 & 0.579 & 0.689 \\
Bicubic & 0.691 & 0.410 & 0.353 & 0.734 & 0.327 & 0.293 \\
SLERP & 0.673 & 0.323 & 0.325 & 0.679 & 0.301 & 0.311 \\
Symm-SLERP & 0.613 & 0.374 & 0.381 & 0.566 & 0.382 & 0.406 \\
Atindama inpainting~\cite{atindama2023restoration} & 0.813 & 0.355 & 0.245 & 0.792 & 0.358 & 0.263 \\
\qrbsaadapted~\cite{jangid2024qrbsa} & 0.335 & 0.665 & 0.665 & \textbf{0.076} & 0.664 & 0.773 \\
QEDSR~\cite{lim2017edsr} & 0.080 & 0.796 & 0.853 & 0.104 & 0.669 & 0.766 \\
EDSR~\cite{lim2017edsr} & 0.083 & 0.791 & 0.849 & 0.147 & 0.652 & 0.739 \\
RCAN~\cite{zhang2018rcan} & 0.070 & 0.810 & 0.866 & 0.476 & 0.499 & 0.511 \\
SAN~\cite{dai2019san} & 0.036 & \textbf{0.843} & \textbf{0.899} & 0.106 & \textbf{0.731} & \textbf{0.804} \\
HAN~\cite{niu2020han} & 0.151 & 0.777 & 0.812 & 0.279 & 0.627 & 0.671 \\
\textbf{SG-SRAN (ours)} & 0.038 & 0.830 & 0.891 & 0.093 & 0.687 & 0.782 \\
\botrule
\end{tabular}
}
}
\end{table*}

\clearpage

\begin{table*}[p]
\caption{\textbf{Zero-shot boundary-window composition analysis.}
The same $3\times3$ HR-boundary-window composition diagnostic as Extended Data Table~\ref{tab:boundary_composition_indist} is applied to the two headline out-of-distribution targets. Spurious rate is lower better; recall and Composition F1 are higher better.}\label{tab:zero_shot_boundary_composition}
\centering
{\scriptsize
\setlength{\tabcolsep}{3.2pt}
\resizebox{\textwidth}{!}{%
\begin{tabular}{lcccccc}
\toprule
Method &
\multicolumn{3}{c}{IN718 $\to$ CoNi} &
\multicolumn{3}{c}{\TiDataset{} $\to$ Ti-Al} \\
\cmidrule(lr){2-4}\cmidrule(lr){5-7}
& Spurious $\downarrow$ & Recall $\uparrow$ & F1 $\uparrow$ &
Spurious $\downarrow$ & Recall $\uparrow$ & F1 $\uparrow$ \\
\midrule
Nearest & 0.054 & 0.609 & 0.741 & 0.014 & 0.699 & 0.818 \\
Bicubic & 0.620 & 0.393 & 0.387 & 0.626 & 0.419 & 0.395 \\
SLERP & 0.598 & 0.332 & 0.364 & 0.569 & 0.393 & 0.411 \\
Symm-SLERP & 0.554 & 0.364 & 0.401 & 0.510 & 0.437 & 0.462 \\
Atindama inpainting~\cite{atindama2023restoration} & 0.813 & 0.335 & 0.240 & 0.820 & 0.374 & 0.243 \\
\qrbsaadapted~\cite{jangid2024qrbsa} & 0.323 & 0.600 & 0.636 & 0.007 & 0.769 & 0.867 \\
QEDSR~\cite{lim2017edsr} & 0.090 & 0.716 & 0.801 & 0.049 & 0.769 & 0.850 \\
EDSR~\cite{lim2017edsr} & 0.090 & 0.711 & 0.798 & 0.092 & 0.753 & 0.823 \\
RCAN~\cite{zhang2018rcan} & 0.091 & 0.721 & 0.804 & 0.578 & 0.496 & 0.456 \\
SAN~\cite{dai2019san} & \textbf{0.045} & \textbf{0.756} & \textbf{0.844} & 0.046 & \textbf{0.852} & 0.900 \\
HAN~\cite{niu2020han} & 0.160 & 0.693 & 0.760 & 0.283 & 0.705 & 0.711 \\
\textbf{SG-SRAN (ours)} & 0.054 & 0.745 & 0.834 & \textbf{0.005} & 0.847 & \textbf{0.915} \\
\botrule
\end{tabular}
}
}
\end{table*}

\end{document}